\RequirePackage{fix-cm}
\documentclass[twocolumn]{svjour3}          
\smartqed  

\usepackage{paralist}
\usepackage{colortbl}
\usepackage{adjustbox}
\usepackage{booktabs}
\definecolor{green}{rgb}{1,0,0}
\usepackage{bbding}
\usepackage{makecell}
\usepackage{subcaption}
\usepackage{graphicx}
\usepackage{amsmath}
\usepackage{multirow}
\usepackage{pifont}
\usepackage{graphicx}
\usepackage{amsmath}
\usepackage{amssymb}
\usepackage{cite}

\def\eg{\emph{e.g.}}
\def\ie{\emph{i.e.}}

\def\0{{\bf 0}}
\def\1{{\bf 1}}

\def\st{\mbox{s.t. }}

\usepackage{listings} 
\usepackage{tcolorbox}

\newcommand{\lstfont}[1]{\color{#1}\ttfamily}

\usepackage{xspace}
\usepackage[colorlinks,linkcolor=red,anchorcolor=blue,citecolor=purple,CJKbookmarks=True]{hyperref}

\begin{document}

\title{Unsupervised Recognition of Unknown Objects for Open-World Object Detection
}

\author{Ruohuan Fang,
        Guansong Pang,
        Lei Zhou,
        Xiao Bai,
        Jin Zheng
}
\authorrunning{R. Fang, G. Pang, L. Zhou, B. Bai,  J. Zheng} 

\institute{
Ruohuan Fang, Xiao Bai, Jin Zheng\at
              School of Computer Science and Engineering, State Key Laboratory of Software Development Environment, Jiangxi Research Institute, Beihang University \\
              \email{ruohuanfang@gmail.com, baixiao@buaa.edu.cn, jinzheng@buaa.edu.cn}
           \and
           Guansong Pang \at
           School of Computing and Information Systems, Singapore Management University \\
           \email{gspang@smu.edu.sg} 
           \and
           Lei Zhou \at School of Computer Science and Artificial Intelligence, Wuhan University of Technology
               \\
              \email{leizhou@whut.edu.cn}  
            \and
            Corresponding Authors: G. Pang and X. Bai.
}

\date{\today}

\maketitle

\begin{abstract}
Open-World Object Detection (OWOD) extends object detection problem to a realistic and dynamic scenario, where a detection model is required to be capable of detecting both known and unknown objects and incrementally learning newly introduced knowledge. Current OWOD models, such as ORE and OW-DETR, focus on pseudo-labeling regions with high objectness scores as unknowns, whose performance relies heavily on the supervision of known objects. While they can detect the unknowns that exhibit similar features to the known objects, they suffer from a severe label bias problem that they tend to detect all regions (including unknown object regions) that are dissimilar to the known objects as part of the background. To eliminate the label bias, this paper proposes a novel approach that learns an unsupervised discriminative model to recognize true unknown objects from raw pseudo labels generated by unsupervised region proposal methods. The resulting model can be further refined by a classification-free self-training method which iteratively extends pseudo unknown objects to the unlabeled regions. Experimental results show that our method 1) significantly outperforms the prior SOTA in detecting unknown objects while maintaining competitive performance of detecting known object classes on the MS COCO dataset, and 2) achieves better generalization ability on the LVIS and Objects365 datasets. Code is available at \url{https://github.com/frh23333/mepu-owod}.


\end{abstract}

\section{Introduction}\label{sec:intro}

Recent years have witnessed tremendous progress in deep learning-based object detection ~\cite{ren2015faster, redmon2016you, zhou2019objects, lin2017focal, zhu2020deformable}. However, traditional object detection models typically adopt a closed-world setting, meaning that they only consider manually annotated known objects and ignore all other unlabeled objects. But there are scenarios where the recognition of unknown objects is critical. For instance, an autonomous car or robot needs to detect unknown obstacles to avoid collisions and ensure safety.

Dhamija et al.~\cite{dhamija2020overlooked} is an early work that explores the problem of Open-Set Object Detection (OSOD), where the detector is trained using known object labels but is required to identify unknown objects during testing. Joseph et al.~\cite{joseph2021towards} further extend the OSOD task to a more dynamic scenario, Open-World Object Detection (OWOD), where the model is required to recognize both known and unknown objects and can be incrementally trained with newly introduced knowledge. 
Previous methods \cite{joseph2021towards, gupta2022ow, zohar2023prob, ma2023cat, han2022expanding} tackle the OWOD problem by pseudo-labeling regions that do not overlap with known objects and have high objectness scores as unknowns. Such objectness scores indicate the probability of the regions belonging to the foreground, which can be obtained via a class-agnostic detector (\eg, RPN\cite{ren2015faster}) trained using known object labels. These methods can successfully detect unknown objects that exhibit similar features to the known objects. However, they suffer from a severe \textit{label bias problem of the known classes}, \ie, they tend to detect all regions (including unknown object regions) that are dissimilar to the known objects as part of the background.

Some previous methods \cite{dong2022open, ma2023cat} have explored using unsupervised region proposal generation methods~\cite{zitnick2014edge, uijlings2013selective, krahenbuhl2014geodesic, wang2022freesolo, bar2022detreg} to improve the generalization ability of OWOD models. Such unsupervised region proposals are usually generated using hand-crafted low-level features (\eg, color, texture, shape, and contour). Such region proposals provide the prior knowledge and physical constraints about the regions where unknown objects may exist. However, they still require objectness-based pseudo labels to calibrate such region proposals produced by unsupervised methods. Therefore, the label bias problem that obstructs the detection of unknown objects remains unsolved in the OWOD task.  

To tackle the aforementioned label bias problem, we propose a novel OWOD framework, named MEPU, to \underline{m}odel and \underline{e}xtend \underline{p}seudo \underline{u}nknown objects generated by unsupervised region proposal generators. MEPU first learns an unsupervised discriminative model to recognize unknown objects from a large set of raw pseudo labels generated by the unsupervised proposal generators. The key insight here is that the common background regions (\eg, sky, sea, grassland, and white wall) frequently appear in different images and usually have repeated and simple low-level patterns (\eg, color, texture). In contrast, the foreground regions have much more diverse visual patterns since there are various object categories. As a result, the background and foreground regions respectively form two different distributions from the perspective of the feature frequency domain. 
Inspired by the reconstruction-based out-of-distribution (OOD) detection models \cite{denouden2018improving, zhou2022rethinking, jiang2023read, osada2023out}, we find that the encoder-decoder framework (\eg, autoencoders) is an ideal tool for modeling such regional features with different appearance frequencies. Therefore, we implement the unsupervised modeling of foreground and background using a new module, called Reconstruction Error-based Weibull Model (REW), which models and identifies the unknown objects by a Weibull modeling of the reconstruction errors of the regions. Based on the modeled Weibull distributions of foreground and background regions, all pseudo labels are assigned with soft labels to estimate the likelihood of being true unknown objects. Our REW module can effectively address the label bias problem since we leverage the feature frequencies, rather than similarity to known objects like the objectness scores, to recognize the unknown objects. Note that the reconstruction-based OOD detectors fail to work in OWOD, since they are difficult to adaptively determine a decision threshold for the unknown objects per training image (see Sec. \ref{sec:ablation}). We address this issue via the Weibull modeling in the REW module.

MEPU further trains a REW-enhanced Object Localization Network (ROLNet)
for extending pseudo unknown labels to the unlabeled regions. More specifically,
the soft labels yielded from REW are incorporated in the training loss of a classification-free detector Object Localization Network (OLN)~\cite{kim2022learning}, so that the pseudo labels of unknown objects can be effectively and accurately extended in each round of self-training, which further alleviates the impact of label bias problem. 

\begin{figure}
\centering
\resizebox{1\linewidth}{!}{
    \includegraphics[width=.9\textwidth]{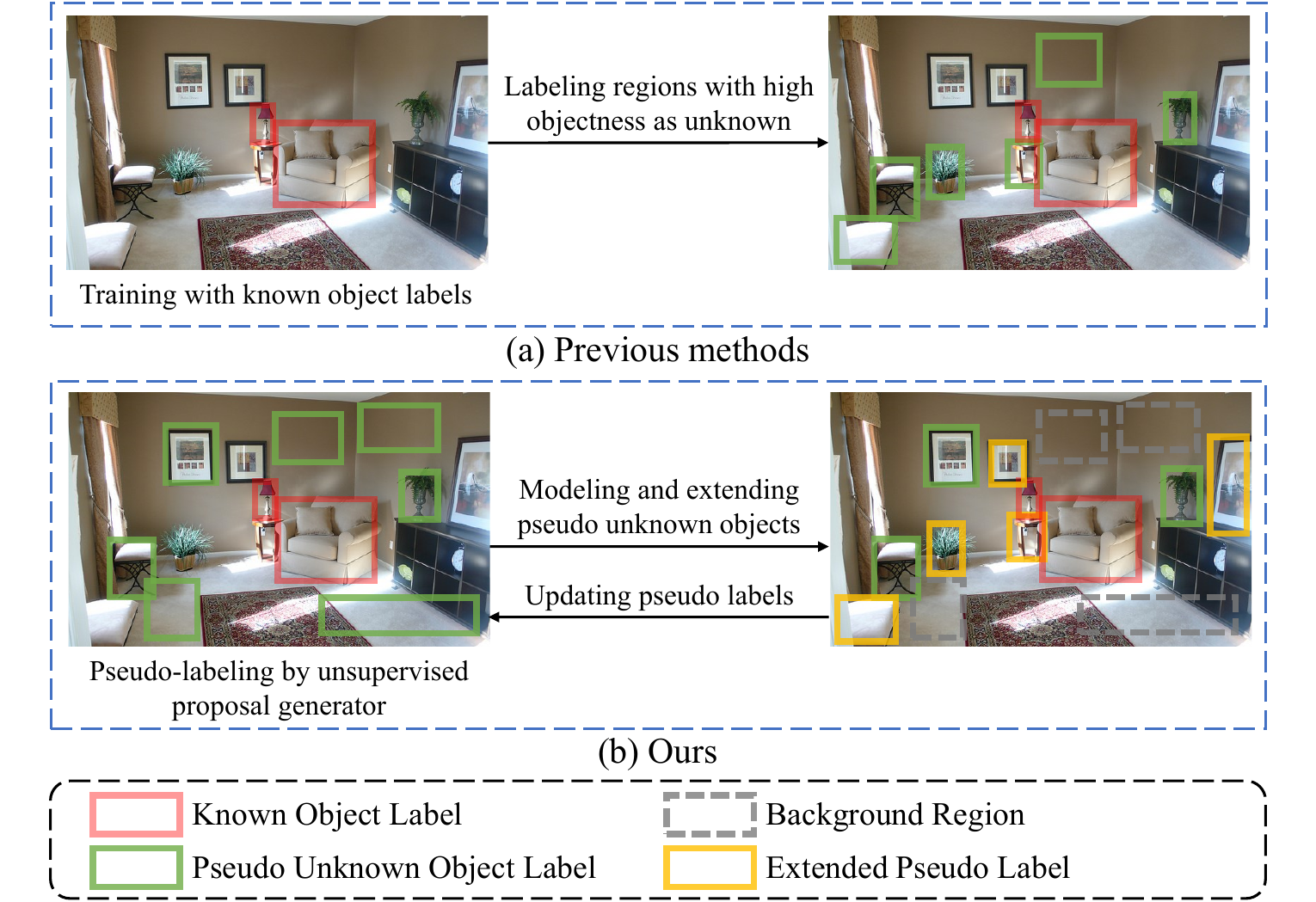}}
    \caption{\textbf{Previous methods vs. ours.} (a) Previous OWOD methods~\cite{joseph2021towards, gupta2022ow} suffer from a severe label bias problem that they tend detect all regions (including unknown object regions) that are dissimilar to the known objects as part of the background, while (b) our approach eliminates this label bias problem by modeling and extending pseudo unknown objects in an unsupervised manner, achieving significantly improved ability to detect unknown objects, without affecting the performance of known object detection. 
    }
\label{fig:intro}
\end{figure}

Our contributions can be summarized as follows:
\begin{compactitem}
    \item We propose a novel approach named MEPU that decouples the unknown object recognition problem into two sub-problems: unsupervised modeling of unknown objects in the raw pseudo labels generated by unsupervised region generators, and extending new unknown objects that are not covered in these pseudo labels. 
    \item We introduce a novel module, namely Reconstruction Error-based Weibull Model (REW), that utilizes the prior knowledge of object occurrence frequency via Weibull modeling to have an effective unsupervised recognition of unknown objects and tackle the label bias problem in OWOD task. 
    \item We further train a REW-enhanced Object Localization Network (ROLNet) that synthesizes soft labels yielded from REW and classification-free self training to effectively and accurately extend the coverage of pseudo labels. 
    \item Experiments show that our proposed method MEPU 1) significantly outperforms the prior SOTA in detecting unknown objects (\eg, having up to $\sim$13.9 increase in unknown recall rate) while maintaining competitive performance of detecting known object classes on the MS COCO dataset, and 2) achieves better generalization ability on the LVIS and Objects365 datasets. 
\end{compactitem}

\section{Related Work}
\noindent\textbf{Open-Set Object Detection.}
Dhamija et al.~\cite{dhamija2020overlooked} is the first work to explicitly explore the OSOD problem, where the detector is required to identify unknown objects during testing. ~\cite{miller2019evaluating, miller2018dropout} use Dropout Sampling (DS) to measure the uncertainty of the object detector and reject the unknown objects. 
VOS~\cite{du2022vos} synthesizes virtual outliers that can regularize the model’s decision boundary during training. OpenDet~\cite{han2022expanding} identifies unknown objects by separating high/low-density regions in the latent space, and it uses Contrastive Feature Learner (CFL) and Unknown Probability Learner (UPL) to achieve this goal. 

Some other works ~\cite{kim2022learning, ren2015faster, saito2022learning, wang2022open, qi2022open} focus on generating class-agnostic region proposals with higher generalization ability for objects of unknown categories. Kim et al.~\cite{kim2022learning} propose an Object Localization Network (OLN) which replaces the classification heads of Faster-RCNN~\cite{ren2015faster} with localization quality prediction heads in order that the detector will not overfit the known classes and depress the scores of unknown objects. ~\cite{saito2022learning} propose a new augmentation method BackErase that pastes known objects on a background image sampled from a small region of the original image, so that potential unknown objects will not be suppressed as negatives. ~\cite{wang2022open} propose the Generic Grouping Networks (GGNs) that learn the Pairwise Affinities (PA) of pixels and use the PA predictions to construct pseudo labels which generalize well to unknown objects. \cite{qi2022open} formulate the entity segmentation task which requires segmenting all visual entities within an image, including things (object) and stuff. They devise an entity segmentation framework based on the unified center-based representation and propose the global kernel bank and overlap suppression module to generate high-quality class-agnostic segmentation masks. 

\noindent\textbf{Open-World Object Detection.} Joseph et al.~\cite{joseph2021towards} extend the OSOD task to a more dynamic scenario and formulate the OWOD problem, where the model is required to recognize both known and unknown objects and can be incrementally trained with newly introduced knowledge.  
They propose the ORE model, which uses the objectness scores of RPN for pseudo-labeling unknown objects and adopts an energy-based classifier to separate the known and unknown classes.
OW-DETR~\cite{gupta2022ow} adopts Deformable DETR (detection transformer) as the base detector and uses its attention maps obtained from the intermediate features as scores to assign pseudo labels of unknown classes. 
PROB ~\cite{zohar2023prob} further incorporates
a probabilistic objectness head into the standard Deformable DETR model, which iteratively estimates the objectness probability distribution and maximizes the likelihood of known objects to learn more general features for both known and unknown objects. CAT ~\cite{ma2023cat} decouples the detection process via the shared decoder in the cascade decoding way and adopts the self-adaptive pseudo-labelling mechanism which combines the model-driven and input-driven pseudo-labels and self-adaptively to generate robust pseudo-labels for unknown objects. 
Wu et al.~\cite{wu2022uc} propose Unknown-Classified Open World Object Detection (UC-OWOD) which requires the model to classify unknown objects into different unknown classes. They design the similarity-based unknown classification (SUC) to detect unknown objects as different classes, and the unknown clustering refinement (UCR) to refine the classification of unknown objects. 

Although previous methods have made remarkable progress in the recognition of unknown objects for the OWOD task, they still utilize the objectness scores for unknown object pseudo-labeling, which depends on the supervision of known object labels. Therefore, they suffer from the aforementioned label bias problem that limits their recognition ability of unknown objects that are semantically irrelevant to known ones. 
Instead, our model enables the unsupervised discriminative recognition of unknown objects, which effectively addresses the label bias problem and improves the OWOD model's detection performance of unknown objects.

\noindent\textbf{Unsupervised Region Proposal Generation Methods.} 
Before the deep learning era, many works~\cite{alexe2010object, carreira2011cpmc, zitnick2014edge, uijlings2013selective, pont2016multiscale, krahenbuhl2014geodesic} 
focus on generating region proposals as object candidates based on hand-crafted low-level features (\eg, color, texture, and contour). 
Selective Search~\cite{uijlings2013selective} greedily merges superpixels to generate proposals. 
EdgeBoxes~\cite{zitnick2014edge} produces proposals by scoring bounding boxes based on the number of edge contours. 
But they are later replaced by deep learning-based supervised methods~\cite{ren2015faster} due to low precision and high time cost. Geodesic Object Proposal~\cite{krahenbuhl2014geodesic} identifies a set of candidate objects based on critical level sets in geodesic distance transforms computed for seeds placed in the image. 
Recently, some works~\cite{wang2022freesolo, bar2022detreg} have explored generating deep network-based region proposals in an unsupervised manner. 
FreeSOLO~\cite{wang2022freesolo} learns class-agnostic instance segmentation without any manual annotations and its proposal quality significantly outperforms previous unsupervised methods. Detreg\cite{bar2022detreg} is trained on ImageNet\cite{deng2009imagenet}  using Selective Search to provide pseudo ground truth labels. 

Although these unsupervised region proposal generation methods can not accurately locate each object, they provide additional knowledge and geometric constraints about the regions where unknown objects may exist. In our proposed method, we leverage those unsupervised region proposal generators to produce raw pseudo-labels of unknown objects. 

\noindent\textbf{Reconstruction-based OOD Detection.} OOD Detection aims to detect and reject testing samples that do not belong to the distribution the model has been trained on. The core idea of reconstruction-based methods is that the encoder-decoder framework trained on the ID (In Distribution) data has larger reconstruction errors for OOD data, so we can separate ID and OOD samples based on their reconstruction errors during inference. ~\cite{denouden2018improving} incorporates the Mahalanobis distance in latent space to better capture OOD samples that lie far from ID samples in latent space but near the latent dimension manifold of the model. ~\cite{zhou2022rethinking} formulates the essence of the reconstruction-based approach as a quadruplet domain translation with an intrinsic bias to only query for a proxy of conditional data uncertainty. Accordingly, they adopt strategies including semantic reconstruction, data certainty decomposition, and normalized L2 distance to substantially improve OOD detection performance. READ~\cite{jiang2023read} incorporates auto-encoder into classifier-based OOD detection model by transforming the reconstruction error of raw pixels to the latent space of the classifier.

Our REW module is inspired by the reconstruction-based method in OOD detection. The key difference is that the auto-encoder in our method reconstructs all regional features including frequently-appeared background regions and rarely-appeared foreground regions, so that their reconstruction errors can be used to model the probability distributions for background-foreground recognition; whereas the OOD detection methods are focused on reconstructing image-level ID samples. Moreover, REW learns a soft label indicating the probability of being a true object for each pseudo unknown object via our Weibull modeling. While the reconstruction-based OOD detection models simply utilize reconstruction errors as the OOD scores to reject testing samples whose scores that are higher than a pre-defined user threshold. In our problem setting, such a hard thresholding strategy cannot be adaptive to each training image, leading to high false negative errors, \ie, true unknown objects are filtered out by the hard threshold as background regions.

\begin{figure*}[h]
    \centering
        \centering
        \includegraphics[width=0.98\textwidth]{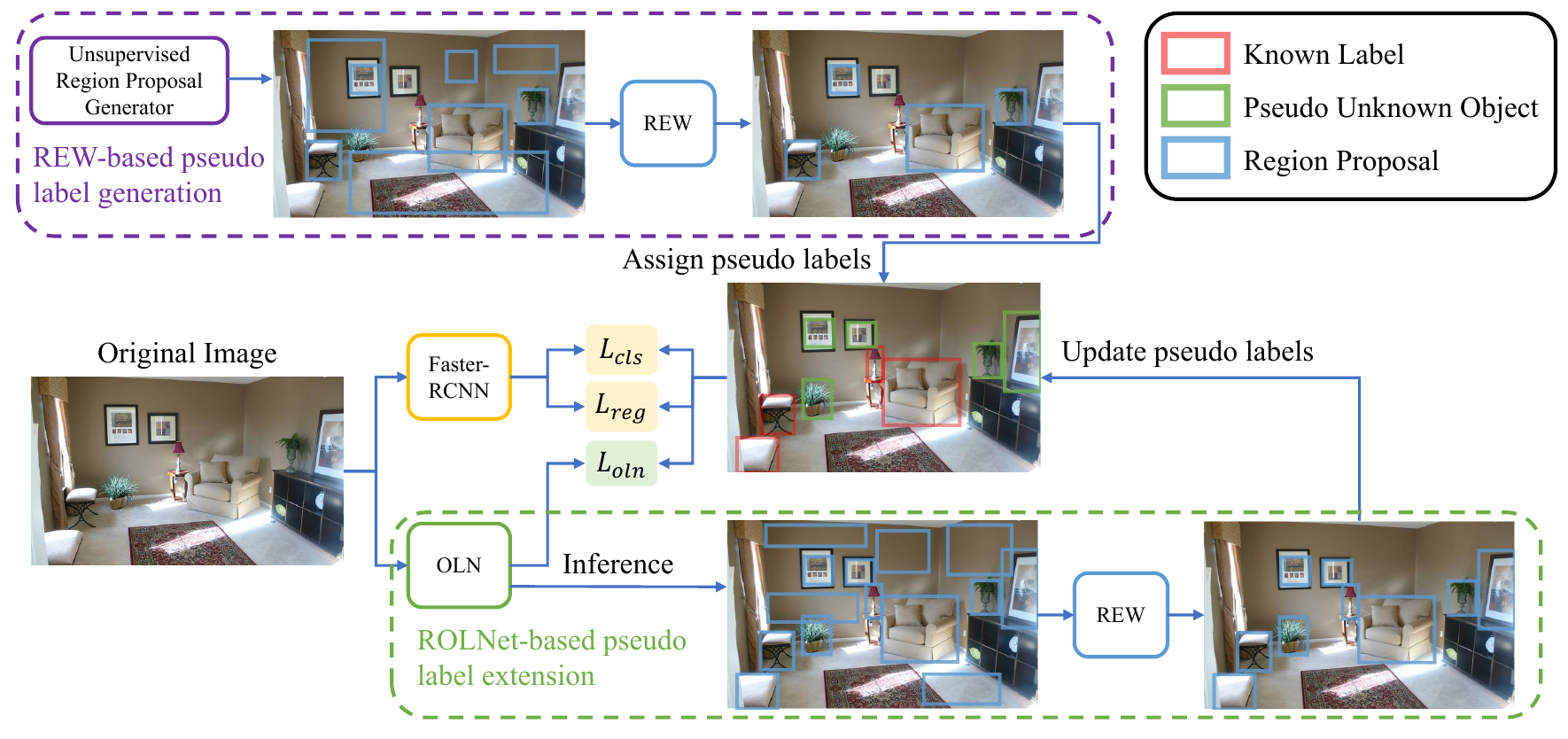}
        \caption{\textbf{Our proposed approach.} It consists of two main components, including \textit{\underline{R}econstruction \underline{E}rror-based \underline{W}eibull Model} (REW) and \textit{\underline{R}EW-enhanced \underline{O}bject \underline{L}ocalization \underline{Net}work} (ROLNet). REW is designed for modeling unknown objects hidden in the pseudo labels generated by unsupervised proposal generators, while ROLNet is employed to iteratively extend new unknown objects based on the likelihood scores produced by REW. }
        \label{fig:model}
\end{figure*}

\section{Proposed Method}

\subsection{Problem definition}\label{sec:def}
\noindent In OWOD, we are given a streaming set of object detection tasks  $\mathcal{T}^t=\{T_1,T_2,\cdots\}$. For each task $T_t$, a set of known classes $\mathcal{K}^t=\{C_t^1,C_t^2,\cdots,C_t^K\}$ and a set of unknown classes $\mathcal{U}^t=\{C_t^{K + 1},C_t ^{K+ 2},\cdots\}$ are presented. The training dataset $\mathcal{D}^t=\{\mathcal{I}^t,\mathcal{Y}^t\}$ contains N images $\mathcal{I}^t=\{I_1,I_2,\cdots,I_N\}$ with the corresponding known class labels $\mathcal{Y}^t=\{Y_1,Y_2,\cdots,Y_N\}$. Each $Y_i=\{y_1,y_2,\cdots,y_K\}$ denotes the annotations of K instances in one image. The k-th instance label is $y_k=[l_k,x_k,y_k,w_k,h_k]$, where $l_k\in \mathcal{K}^t$ denotes the class label, and $[x_k,y_k,w_k,h_k]$ denotes the coordinates, width, and height of the bounding box.

OWOD has two main problems, including open-set object detection and incremental object detection~\cite{shmelkov2017incremental, hao2019end, kj2021incremental, yang2022continual}. For the former problem, the goal is to train an object detector $\mathcal{M}^t$ at time step $t$ with $\mathcal{D}^t$ that contain only the annotations of known object classes. At testing time, the model should be able to detect known objects by classifying them into one of the $\mathcal{K}_t$ classes, while identifying objects beyond known classes by labeling them as `\textit{unknown}'. In the incremental object detection, a new task $T_{t+1}$ at the next time step is given, where a new set of known classes
$\mathcal{K}^{t+1}=\{C_{t + 1}^1, C_{t + 1}^2,\cdots,C_{t+1}^M\}$ 
is presented, then the goal is to adapt the model $\mathcal{M}^t$ to the new knowledge in $\mathcal{K}^{t+1}$ and obtain a new model $\mathcal{M}^{t+1}$, instead of retraining from scratch on the whole dataset. $\mathcal{M}^{t+1}$ is required to be capable of correctly identifying all the newly added known classes, as well as the previously known and unknown classes, \ie, without catastrophic forgetting of the previously learned knowledge \cite{goodfellow2013empirical, french1999catastrophic, mccloskey1989catastrophic}.

\subsection{Overview of The Proposed Approach} \label{sec:overview}

\noindent To address the label bias towards the known classes, our proposed approach MEPU aims to first have unsupervised modeling of unknown objects presented in the raw pseudo labels generated by unsupervised region proposal generators, and then extend to new unknown objects that are not covered in these pseudo labels. 
To this end, our approach introduces two main modules, consisting of \textit{\underline{R}econstruction \underline{E}rror-based \underline{W}eibull Model} (REW) and \textit{\underline{R}EW-enhanced \underline{O}bject \underline{L}ocalization \underline{Net}work} (ROLNet), as illustrated in Fig.~\ref{fig:model}. 
REW can be considered as an unsupervised unknown object identification model that is trained to recognize various types of unknown objects from common background regions in an unsupervised manner and assign all pseudo labels with soft labels to estimate the likelihood of being true unknown objects. While ROLNet leverages the REW scores and classification-free OLN in the self-training procedure to effectively extend the set of unknown objects that exist in the unlabeled regions. Faster-RCNN \cite{ren2015faster} is used as the base detector in our approach. Below we introduce these two modules in detail. 

\subsection{REW: Reconstruction Error-based Weibull Model for Unsupervised Unknown Object Identification} \label{sec:rew}
Existing unsupervised region proposal methods can generate massive regions that may contain various types of unknown objects, and thus these region proposals that do not overlap with known instances can be used as pseudo labels for unknown objects. However, these raw unknown pseudo labels may also include many non-object bounding boxes on the background. The key challenge is to distinguish the true unknown objects from these raw pseudo labels while preventing the label bias problem. 
The proposed REW module (Reconstruction Error-based Weibull Model) is specifically designed for learning an unsupervised unknown object identification model to tackle this challenge. 

REW is motivated by the observation that the common background regions (\eg, sky, sea, grassland, and white wall) frequently appear in different images and usually share repeated and simple low-level patterns (\eg, color, texture). In contrast, the foreground regions have much more diverse visual patterns since there are various object categories. As a result, the background and foreground regions respectively form two different distributions from the perspective of the feature frequency domain. Therefore, REW is devised to learn the frequency information by a data reconstruction-based auto-encoder and model the two different distributions of the reconstruction errors by a prior probability distribution. Since Weibull distribution is superior in fitting a wide range of distribution shapes, it is used as the prior model in REW. 

\begin{figure*}[h]
    \centering
        \centering
        \includegraphics[width=1.0\textwidth]{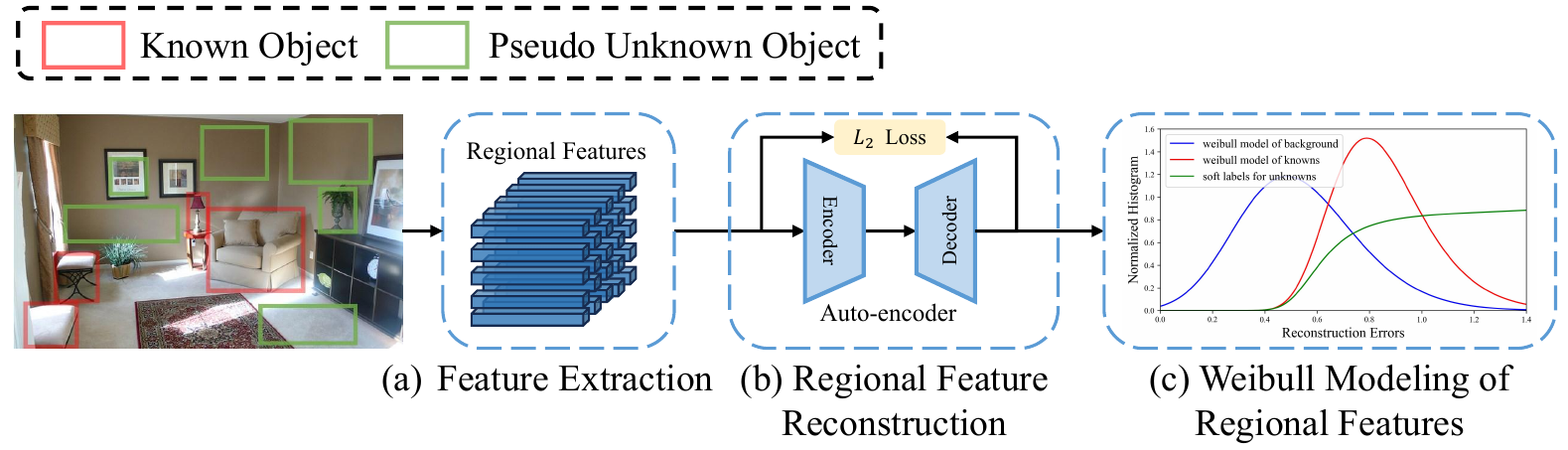}
        \caption{\small \textbf{The pipeline of our reconstruction error-based Weibull modeling (REW) module.} REW first extracts the regional features from the backbone and then feeds all feature vectors to train an auto-encoder by minimizing the errors of reconstructing the regional features. 
        The distribution of the $\ell_2$ distance-based reconstruction errors for the known objects (red boxes) and background regions are modeled using a Weibull model. The pseudo unknown objects (green boxes) would be assigned with soft labels using Eq. (\ref{equation:soft_label}) to provide knowledge of true unknown objects for training generalized open-world object detectors that are able to detect different types of unknown objects while preventing label bias problem.
        }
       \label{fig::rew}
\end{figure*}

\subsubsection{Reconstruction Error-based Weibull Modeling of Foreground and Background Regions} \label{sec:weibull}
As demonstrated in Fig.~\ref{fig::rew}, given an input image $\mathbf{I} \in \mathbb{R}^{H_I \times W_I \times 3}$, we first use a backbone network to extract the feature map $\mathbf{F} \in \mathbb{R}^{H_F \times W_F \times C}$. The feature vector of each pixel in the feature map represents the features of the regions within the receptive field. Therefore, we utilize the auto-encoder to reconstruct such regional features. The encoder and decoder of the auto-encoder are built as multiple convolutional layers with ${1 \times 1}$ convolutional kernels, noted as $Enc()$ and $Dec()$ respectively. The encoder first maps the feature map $F$ into a latent space $\mathbf{F_{latent}} \in \mathbb{R}^{H_F \times W_F \times C_{latent}}$ with fewer dimensions (channels). Then, the decoder reconstructs the latent feature into the original dimensions, obtaining the reconstructed features $F_{rec} \in \mathbb{R}^{H_F \times W_F \times C}$. We measure the per-pixel reconstruction error by an $\ell_2$ distance and utilize it as the training loss for the auto-encoder. The process can be formulated as follows: 
\begin{equation}
F_{rec} = Dec(Enc(F)) 
\end{equation}

\begin{equation} \label{equ:rew}
L_{rew} = \frac{1}{H_F \times W_F}\sum_{i=1}^{H_F} \sum_{j=1}^{W_F}L_2(F_{rec}[i,j], F[i,j]),
\end{equation}
where $F_{rec}[i,j]$ and $F[i,j]$ denote the regional features with $C$ dimensions in the location $[i,j]$; $L_2$ means the $\ell_2$-norm loss. 

The regional feature of each pixel represents the features of the anchor in the corresponding location, so we can assign each pixel with a foreground/background label based on its corresponding anchor box (\eg, in RPN, an anchor box belongs to the foreground when it overlaps any object with IoU larger than 0.7). As discussed in Sec.\ref{sec:intro}, background regions often share frequently-occurred features, making them better reconstructed and resulting in smaller reconstruction errors compared to the infrequent features of various foreground object regions.

\begin{figure}[t]
    \centering
        \centering
        \includegraphics[width=.45\textwidth]{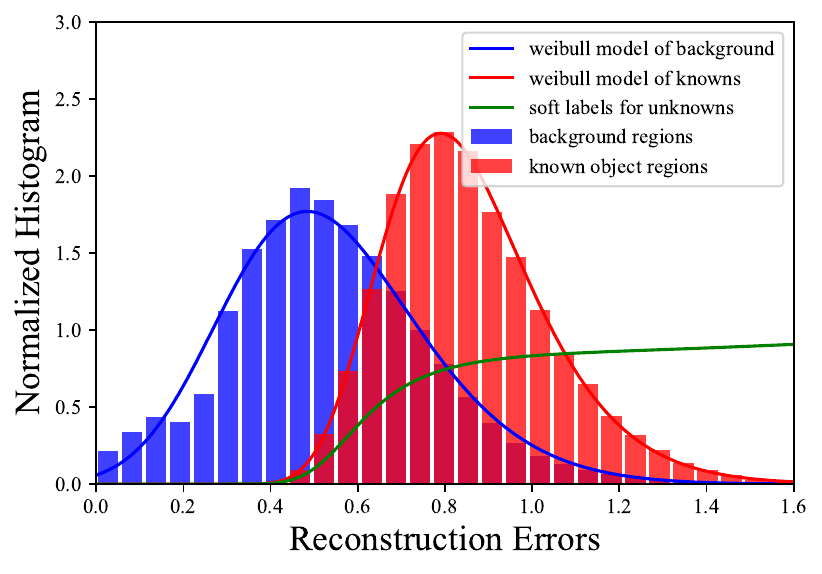}
        \caption{\small \textbf{Visualization of reconstruction error distributions on the MS COCO dataset.} 
        The known and background regions are randomly sampled from the training images of MS COCO.} 
        \label{fig:dist_rew}
    \end{figure}

When the auto-encoder is trained to convergence, we obtain the reconstruction error map $\mathbf{E} \in \mathbb{R}^{H_F \times W_F \times 1}$ via calculating per-pixel $\ell_2$ distance, \ie, $E[i,j] =\\ L_2(F_{rec}[i,j], F[i,j])$. By randomly sampling pixels from known objects and background regions, we collect a set of reconstruction errors (noted as $\mathcal{E}_{kn}$ and $\mathcal{E}_{bg}$ respectively). Fig.~\ref{fig:dist_rew} visualizes the histogram of randomly sampled reconstruction errors of the annotated known objects regions and background regions from the training set of MS COCO~\cite{lin2014microsoft}. It is evident that the reconstruction errors of the known object regions are generally much larger compared to those of the background regions. Although unknown objects may have different appearances from known objects, we can assume that they have similar low occurrence frequencies and high reconstruction errors, as there are various types of unknown objects. Therefore, we utilize the reconstruction errors from the sampled annotated known objects to estimate the distribution of all foreground regions. 

Since Weibull distribution is superior in fitting a wide range of distribution shapes for many real-world scenarios, it is used as the prior model in REW. The Weibull distributions of known regions and background regions (noted as $f_{kn}$ and $f_{bg}$ respectively) are in the form as follows: 
\begin{equation}
    f(re;a,c) = ac[1-\exp (-re^c)]^{a-1}\exp (-re^c)re^{c-1},
\end{equation}
where $re$ denotes the reconstruction error value of a sampled pixel; $f$ is a probability density function of exponentiated Weibull Distribution and $a$ and $c$ are its shape parameters. The optimal $a$ and $c$ are calculated using Maximum Likelihood Estimation (MLE) based on the sampled reconstruction errors of foreground ($\mathcal{E}_{kn}$) and background regions ($\mathcal{E}_{bg}$).

\subsubsection{Soft Labeling for Unknown Object Recognition} \label{sec:softlabel}
After modeling the distributions of foreground and background regions, we then use the probability function ($f_{kn}$ and $f_{bg}$) to estimate the likelihood of a pseudo unknown object being a true unknown object. Given one pseudo unknown object $p_{unk}\in \mathbb{R}^{4}$ in the image $I$, it may cover multiple pixels in the feature map. Therefore, we calculate its reconstruction error by averaging the values of all corresponding pixels in that location. To be specific, we utilize the RoIAlign\cite{he2017mask} operation which pools the reconstruction errors within $p_{unk}$ into a scalar value, as formulated below: 
\begin{equation}\label{equ::re}
    re(p_{unk}) = RoIAlign(E, p_{unk}),
\end{equation}
where $re(p_{unk})$ is the reconstruction error value of $p_{unk}$; $RoIAlign$ means the RoIAlign operation; $\mathbf{E} \in \mathbb{R}^{H_F \times W_F \times 1}$ denotes the calculated reconstruction error map. Furthermore, we can calculate the soft label, which estimates the likelihood score of a pseudo object being a true unknown object, using the following equation:
\begin{equation}\label{equation:soft_label}
    s(p_{unk}) = (\frac{f_{kn}(re(p_{unk}))}{f_{bg}(re(p_{unk}))+f_{kn}(re(p_{unk}))})^\gamma,
\end{equation}
where $f_{kn}$ and $f_{bg}$ are the Weibull probability density functions of the known object and background regions, respectively, and $\gamma$ is a hyperparameter to scale the value of the likelihood score. All pseudo labels will be discarded when $\gamma \to\infty$ and all pseudo labels will be treated as true objects when $\gamma \to 0$.

\subsubsection{Incorporating the Soft Labels into Object Detection Loss Functions} 
The unknown object likelihood scores $s$ estimated in Eq. (\ref{equation:soft_label}) can then be incorporated into the classification loss ($L_{cls}$) of the detector Faster-RCNN as a weight term to learn to recognize unknown objects while detecting known objects. The modified classification loss equation is as follows:
\vspace{-0.1cm}
\begin{equation}\label{eqn:clsloss}
    L_{cls}
    = \frac{1}{N_{cls}}\sum_{i}w_{r_i}
    L_{CE}(p_{r_i},p_{r_i}^*),
\end{equation}
where $w_{r_i}$ is the loss weight of a region proposal $r_i$ which equals to $s(r_i)$ when $r_i$ belongs to the region of a pseudo unknown object, or otherwise it equals to 1; $p_{r_i}$ and $p_{r_i}^*$ denote the predicted probability and the ground truth of the region proposal $r_i$ respectively and $L_{CE}$ means the cross-entropy loss. 

\subsection{ROLNet: REW-enhanced Object Localization Network for Extending the Set of Unknown Objects\label{sec:clast}}

Although our proposed REW module can effectively identify true unknown objects from the pseudo unknown object labels generated by unsupervised region proposal methods, there are still unknown objects not covered by the pseudo labels.
Current semi-supervised object detection methods~\cite{yang2021interactive, chen2022label, sohn2020simple, zhou2021instant, mi2022active} adopt self-training to extend the pseudo labels to the unlabeled data. However, it cannot work well in our problem setting, because traditional object detectors typically adopt cross-entropy or focal loss as the classification loss for proposal classification, which results in a bias toward the known classes. Since all unlabeled regions are treated as negative samples in the classification loss, minimizing such loss functions would compress the objectness scores of the unlabeled regions as close as zero.

In our ROLNet module, we aim to utilize REW soft labels and the self-training strategy based on classification-free detectors, such as object localization networks (OLN), to address this problem. 
Classification-free detectors aim to estimate the objectness scores by predicting
how well the anchors or proposals are located, such as IoU or centerness with their corresponding ground truth instances. This turns the classification problem into a localization quality prediction problem, where only positive anchors or proposals are sampled for training. So the scores of the unlabeled regions will not be compressed, which helps generalize to the unknown objects missed by the pseudo labels. OLN adopts the same architecture as Faster-RCNN but replaces the classification heads in both stages with the localization quality prediction heads. The localization quality heads are trained with the following loss function:
\begin{equation}
    L_{oln}
    = \frac{1}{N_{oln}}\sum_{i}w_{r_i}L_1(q_{r_i},q_{r_i}^*),
\end{equation}
where $q_{r_i}$ and $q_{r_i}^*$ denote the predicted localization and target scores of the region proposal $r_i$ respectively; $L_{1}$ is an $\ell_1$-norm loss; $w_{r_i}$ is the loss weight term as in Eq.\ref{eqn:clsloss}. The unknown object likelihood score $s(r_{i})$ predicted by REW (\ie, $w_{r_i}$) is incorporated in $L_{oln}$ so that OLN will not be misled by the noisy labels and extend the new pseudo labels more accurately. The positive region proposals in $L_{oln}$ are the proposals that have an IoU larger than 0.3 with the matched ground-truth boxes. By default, centerness and IoU are used as targets of localization scores in the RPN and RoI stages, respectively. 

The overall detection architecture (OLN and Faster-RCNN) is first trained using both known object labels and unknown pseudo labels generated by unsupervised proposal methods. We then apply OLN-based self-training to refine our detection model. The trained OLN is adopted to generate class-agnostic proposals based on its predictions of the localization quality scores. Only the top $P\%$ high-score proposals are selected as new pseudo labels for the unknown objects. After that, the detection model is trained using the updated unknown object labels. This self-training procedure is repeated for $l$ iterations, which can gradually include new unknown objects to further refine the detection model.

\begin{figure}[t]
    \centering
        \centering
        \includegraphics[width=.45\textwidth]{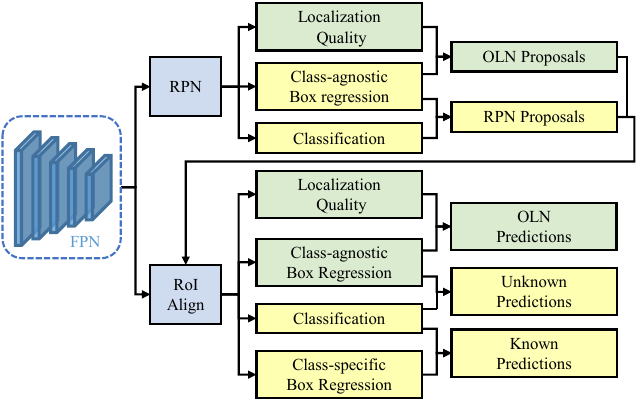}
        \caption{\textbf{The network architecture of our model. }The networks of Faster-RCNN (yellow) 
        and OLN (green) are combined with a unified RPN and RoI Align structure. 
        }
        \label{fig:network}
    \end{figure}

\begin{table}[t]
\renewcommand\arraystretch{1.2}
    \centering
    \resizebox{\columnwidth}{!}{%
    \begin{tabular}{@{}l|cccc@{}} 
    \hline 
        & \textbf{Task 1} & \textbf{Task 2} & \textbf{Task 3} & \textbf{Task 4} \\ \hline
    Semantic split & \begin{tabular}[c]{@{}c@{}}Animals, Person, \\ Vehicles\end{tabular} & \begin{tabular}[c]{@{}c@{}}Appliances, Accessories, \\ Outdoor, Furniture\end{tabular} & \begin{tabular}[c]{@{}c@{}}Sports, \\ Food\end{tabular} & \begin{tabular}[c]{@{}c@{}}Electronic, Indoor, \\ Kitchen\end{tabular} \\\hline
    \# training images & 89,490 & 55,870 & 39,402 & 38,903 \\ \hline
    \# train instances & 421,243 & 163,512 & 114,452 & 160,794 \\ \hline
    \# test images & \multicolumn{4}{c} {4,952} \\ \hline
    \# test instances & \multicolumn{4}{c} {36,781} \\ \hline
        \hline 
    & \textbf{Task 1} & \textbf{Task 2} & \textbf{Task 3} & \textbf{Task 4} \\ \hline
    Semantic split & \begin{tabular}[c]{@{}c@{}}VOC \\ Classes\end{tabular} & \begin{tabular}[c]{@{}c@{}}Outdoor, Accessories, \\ Appliance, Truck\end{tabular} & \begin{tabular}[c]{@{}c@{}}Sports, \\ Food\end{tabular} & \begin{tabular}[c]{@{}c@{}}Electronic, Indoor, \\ Kitchen, Furniture\end{tabular} \\\hline
    \# training images & 16,551 &  45,520 &  39,402 & 40,260 \\ \hline
    \# train instances & 47,223&113,741& 114,452& 138,996 \\ \hline
    \# test images & 4,952& 1,914& 1,642&  1,738 \\ \hline
    \# test instances & 14,976& 4,966& 4,826&6,039 \\ \hline

    \end{tabular}%
    }
    \caption{\textbf{Dataset splits of S-OWODB (superclass-separated OWOD benchmark) (Top) and M-OWODB (superclass-mixed OWOD benchmark) (Bottom). }}
    \label{tab:data_split}
    \end{table}
    
\subsection{Training and Inference\label{sec:train_infer}}
\noindent\textbf{Train. } 
The training process of REW is separate from the object detection model because we have found that features extracted by the self-supervised pre-trained model contain more generalized semantics for recognizing unknown objects, as discussed in \cite{liu2022open, winkens2020contrastive}. In REW, the auto-encoder is trained using regional features extracted by the SoCo backbone~\cite{wei2021aligning} that is pre-trained using object-level contrastive learning. When training the auto-encoder, the pre-trained backbone is frozen as it is only used for feature extraction.  
Subsequently, the trained REW module can calculate likelihood scores for pseudo unknown objects using Eq.\ref{equation:soft_label}. These scores are then utilized in $L_{cls}$ and $L_{oln}$ to train the object detection model. Note that only the known object labels are used in the box regression loss since the pseudo labels are mostly poorly localized. As shown in Fig.~\ref{fig:network}, we combine the network architecture of Faster-RCNN and OLN with a unified RPN and RoI Align structure. For a task $T_i$ with $C_{kn}$ known classes, the Faster-RCNN-based detector is trained as a $(C_{kn} + 2)$-class classifier, where the extra (+2) classes are the `\textit{unknown}' and `\textit{background}' classes. Additionally, we employ localization quality prediction heads in both stages to generate class-agnostic proposals for self-training. 

For the incremental learning procedure, we adopt the approach of exemplar replay~\cite{li2017learning, rebuffi2017icarl, dhar2019learning, chaudhry2018efficient}.
Following the setting of \cite{joseph2021towards, gupta2022ow}, 50 exemplars are stored for each known class, \ie,  a small subset of training images containing 50 objects for each known class. The detector is then finetuned using the stored exemplars to recover the previously known knowledge. 

\noindent\textbf{Inference. }
During inference, the trained Faster-RCNN-based detector is used to classify each bounding box into $(C_{kn} + 2)$ classes. When an unknown object is predicted, we utilize the class-agnostic box regression head for its box localization, because it can produce well-localized unknown object predictions. The inference of known objects is the same as standard Faster-RCNN. 
\begin{figure}[t]
  \centering
  \begin{subfigure}[b]{0.22\textwidth}
    \includegraphics[width=\textwidth]{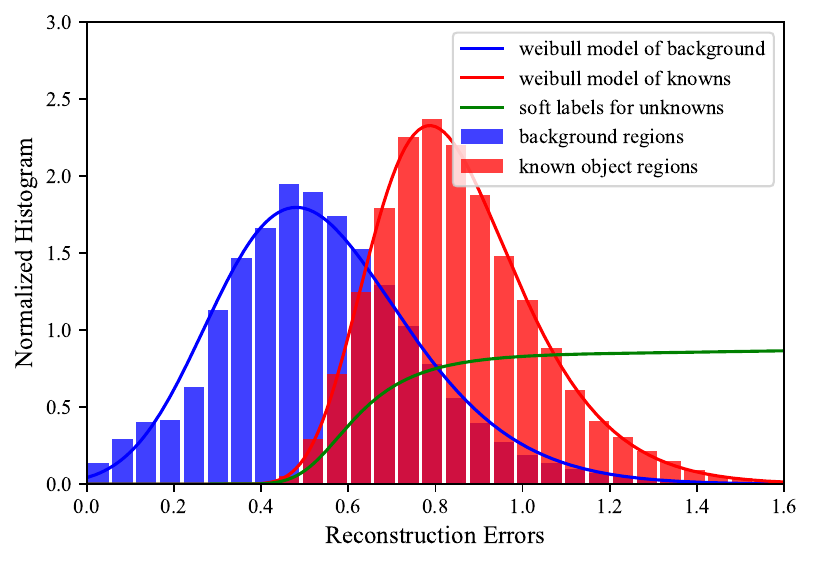}
    \caption{$P_3$}
  \end{subfigure}
  \hfill
  \begin{subfigure}[b]{0.22\textwidth}
    \includegraphics[width=\textwidth]{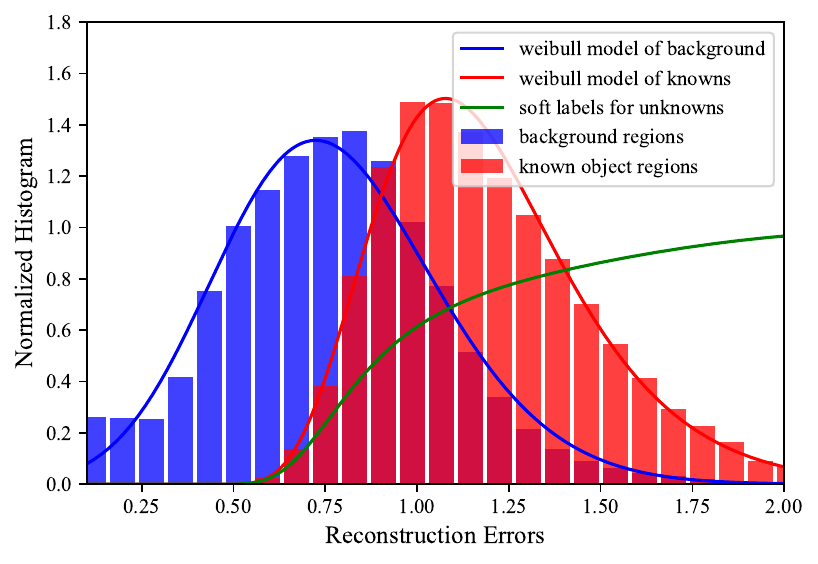}
    \caption{$P_4$}
  \end{subfigure}
  \hfill
  \begin{subfigure}[b]{0.22\textwidth}
    \includegraphics[width=\textwidth]{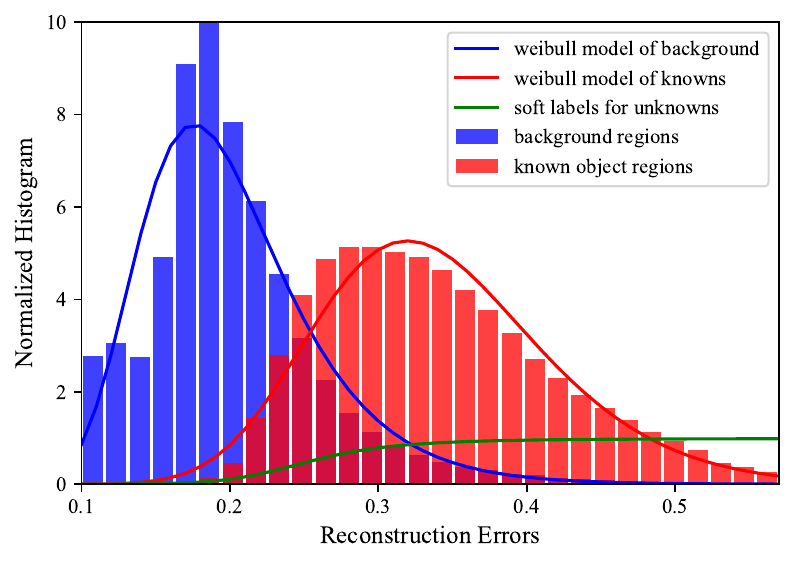}
    \caption{$P_5$}
  \end{subfigure}
  \hfill
  \begin{subfigure}[b]{0.22\textwidth}
    \includegraphics[width=\textwidth]{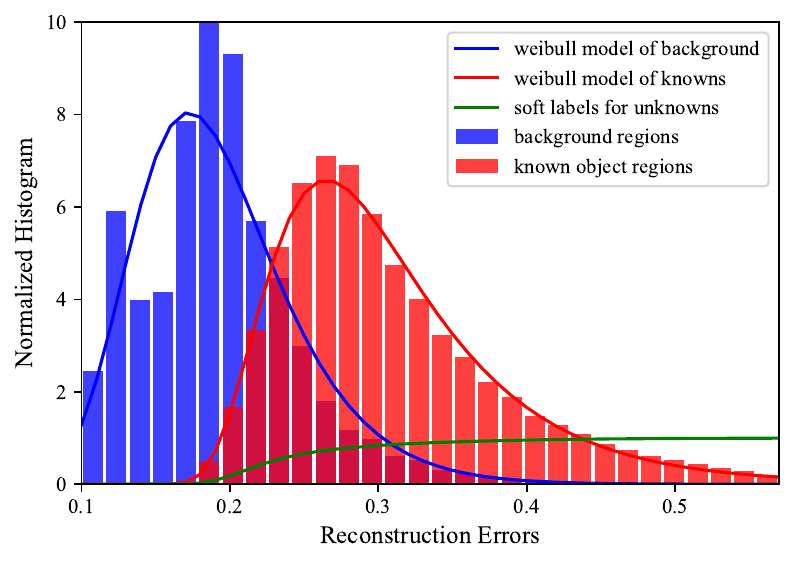}
    \caption{$P_6$}
  \end{subfigure}
  \caption{Visualizations of reconstruction error distributions in different feature levels on S-OWODB Task 1. The Weibull models formed for FPN feature maps $\{P_3,P_4,P_5,P_6\}$ are shown here. }
\end{figure} \label{fig::rew_dists}

\begin{table*}[t]
\centering
\setlength{\tabcolsep}{3pt}
\adjustbox{width=\textwidth}{
\begin{tabular}{@{}l|cc|cccc|cccc|ccc@{}}
\toprule
 \textbf{Task IDs} ($\rightarrow$)& \multicolumn{2}{c|}{\textbf{Task 1}} & \multicolumn{4}{c|}{\textbf{Task 2}} & \multicolumn{4}{c|}{\textbf{Task 3}} & \multicolumn{3}{c}{\textbf{Task 4}} \\ \midrule
 
& {U-Recall} & \multicolumn{1}{c|}{{mAP ($\uparrow$)}} & {U-Recall} & \multicolumn{3}{c|}{{mAP ($\uparrow$)}} & {U-Recall} & \multicolumn{3}{c|}{{mAP ($\uparrow$)}} & \multicolumn{3}{c}{{mAP ($\uparrow$)}}  \\

 & ($\uparrow$) & \begin{tabular}[c]{@{}c}Current \\ known\end{tabular} & ($\uparrow$) & \begin{tabular}[c]{@{}c@{}}Previously\\  known\end{tabular} & \begin{tabular}[c]{@{}c@{}}Current \\ known\end{tabular} & Both & ($\uparrow$) & \begin{tabular}[c]{@{}c@{}}Previously \\ known\end{tabular} & \begin{tabular}[c]{@{}c@{}}Current \\ known\end{tabular} & Both & \begin{tabular}[c]{@{}c@{}}Previously \\ known\end{tabular} & \begin{tabular}[c]{@{}c@{}}Current \\ known\end{tabular} & Both \\ \midrule

Faster-RCNN~\cite{ren2015faster}  & 0.0  & 74.4  & 0.0 & 0.42 & 44.8 & 24.7 & 0.0 & 0.23 & 43.1 & 14.5 & 0.15 & 41.6 & 10.5 \\ 

\begin{tabular}[c]{@{}l@{}}Faster-RCNN\\ \hspace{1em}+ Finetuning\end{tabular} & 0.0  & 74.4  & 0.0 & 68.2 & 42.2 & 54.6  & 0.0 & 50.1 & 38.7 & 46.4 & 43.9 & 35.6 & 41.9 \\ \hline

ORE $-$ EBUI~\cite{joseph2021towards}& 1.5  & 71.4 & 3.9 & 61.0 & 30.9 & 45.6  & 3.6 & 43.1 & 32.2 & 39.5 & 33.6 & 26.3 & 31.8 \\ 

OW-DETR~\cite{gupta2022ow} &  5.7  &  73.1 &  6.2 &  65.0 &  29.0 &  46.0 & 6.9 & 46.7 & 25.7 & 39.7 & 38.2 & 28.1 & 33.1 \\ 

{PROB~\cite{zohar2023prob}} & {17.6}  & {73.5} & {22.3} & {66.3} & {36.0} & {50.4} & {24.8} & {47.8} & {30.4} & 42.0 & {42.6} & {31.7} & {39.9} \\

{CAT~\cite{ma2023cat}} &24.0& 74.2& 23.0& 67.6& 35.5& 50.7& 24.6& \textbf{51.2}& 32.6& 45.0& \textbf{45.4}& \textbf{35.1}& \textbf{42.8} \\

\textbf{MEPU-FS (Ours)} & \textbf{37.9}  & \textbf{74.3} & \textbf{35.8} & \textbf{68.0} & \textbf{41.9} & \textbf{54.3} & \textbf{35.7} & {50.2} & \textbf{38.3} & \textbf{46.2} & {43.7} & {33.7} & {41.2} \\

\textbf{MEPU-SS (Ours) }  & 33.3  & 74.2& {34.2} & {67.5} & 41.0 & {53.6}  & {33.6} & {50.0} & {37.5} & {45.8} & {43.2} & {33.5} & {40.8} \\ 

\bottomrule
\toprule 

Faster-RCNN~\cite{ren2015faster}  & 0.0  & 60.3  & 0.0 & 0.69 & 35.2 & 17.9 & 0.0 & 0.32 & 23.5 & 8.0 & 0.65 & 20.1 & 5.5 \\ 

\begin{tabular}[c]{@{}l@{}}Faster-RCNN\\ \hspace{1em}+ Finetuning\end{tabular} & 0.0  & 60.3  & 0.0 & 57.6 & 34.0 & 45.3  & 0.0 & 43.8 & 22.3 & 36.6 & 35.6 & 19.5 & 31.5 \\ \hline

ORE $-$ EBUI~\cite{joseph2021towards} &  4.9  & 56.0 & 2.9 & 52.7 & 26.0 & 39.4  & 3.9 & 38.2 & 12.7 & 29.7 & 29.6 & 12.4 & 25.3 \\ 

OW-DETR~\cite{gupta2022ow} & 7.5  & 59.2 & 6.2 & 53.6 & 33.5 & 42.9 & 5.7 & 38.3 & 15.8 & 30.8 & 31.4 & 17.1 & 27.8 \\

PROB~\cite{zohar2023prob} & 19.4  & 59.5 & 17.4 & 55.7 & 32.2 & 44.0 & 19.6 &\textbf{43.0} &\textbf{22.2 } &\textbf{36.0 } &\textbf{35.7}  & 18.9 & \textbf{31.5} \\

CAT~\cite{ma2023cat}& 23.7& 60.0& 19.1& 55.5& 32.7& 44.1& 24.4& 42.8& 18.7& 34.8& 34.4& 16.6& 29.9 \\

\textbf{MEPU-FS (Ours) }  &  \textbf{31.6}  & \textbf{60.2 }&  \textbf{30.9} & \textbf{57.3} & 33.3 & \textbf{44.8}  &  \textbf{30.1} & {42.6} & {21.0} & {35.4} & {34.8} & \textbf{19.1} & 30.9 \\ 

\textbf{MEPU-SS (Ours) }  & {30.3}  & {60.0 }&{30.6} &{57.0} &{33.1}& 44.5 & {30.0} & {42.2} & {20.5} & {35.0} & {34.3} & {18.9} & 30.4 \\

\bottomrule

\end{tabular}%
}     
\caption{ \textbf{Comparison to state-of-the-art OWOD models on S-OWODB (top) and M-OWODB (bottom).} We adopt mAP of known classes (Known-mAP) and recall of unknown classes (U-Recall) as the closed-set and open-set evaluation metrics. U-Recall is not available in Task 4 because there are no unknown annotations during inference. Results of our MEPU method using two proposal generation methods, FreeSOLO (MEPU-FS) and Selective-Search (MEPU-SS), are reported. }
\label{tab:sota_owod}
\end{table*}

\section{Experiments}
\subsection{Experimental Settings}

\subsubsection{Datasets.} \label{sec::data}
\noindent\textbf{OWOD Benchmark.} We evaluate our MEPU on two OWOD benchmarks proposed by \cite{gupta2022ow} and \cite{joseph2021towards} respectively. Following \cite{zohar2023prob}, we refer to these two benchmarks as S-OWODB (superclass-separated OWOD benchmark) and M-OWODB (superclass-mixed OWOD benchmark), respectively. As shown in Tab.~\ref{tab:data_split}, the 80 classes of COCO are divided into four groups, and each group of data is treated as the dataset for one of the streaming tasks. During the training procedure of task $T_t$, all classes in $\{T_\tau:\tau \leq t\}$ are known classes, and all classes within $\{T_\tau :\tau > t\}$ are unknown classes. The models are incrementally trained using images with only the known class annotations of each task, instead of using the entire dataset.

\noindent\textbf{Cross Dataset Evaluation.}
The OWOD benchmark focuses on open-set detection within one dataset. Following \cite{kim2022learning, wang2022open}, we also evaluate the detection ability under the cross-dataset setting, using the LVIS v0.5~\cite{gupta2019lvis} and Objects365~\cite{shao2019objects365} datasets. LVIS is a large vocabulary instance segmentation dataset with 1,203 classes in a long-tailed distribution, while Objects365 is a large-scale object detection dataset consisting of 365 object categories. Both datasets include the 80 COCO classes that are treated as known classes and other classes are labeled as unknown. We train our object detector using the training set of MS COCO and evaluate on the testing set of LVIS and Objects365 to measure our generalization performance across datasets.


\subsubsection{Evaluation Metrics.}
We employ mAP (mean Average Precision) as the metric to evaluate the detection performance of known categories. As for unknown classes, we utilize unknown object recall (U-recall) as the primary evaluation metric, following the approaches in \cite{gupta2022ow, ma2023cat, zohar2023prob}. The U-recall metric in our model is calculated using unknown object predictions with scores higher than 0.05. Additionally, we include the results of Recall@K for a more detailed analysis, in accordance with \cite{wang2022open, kim2022learning, saito2022learning}. This metric evaluates the performance based on the top K detected unknown objects, ranked by unknown classification scores. To quantify the extent of misclassification of unknown objects as known ones, we report the Absolute Open-Set Error (A-OSE) and Wilderness Impact (WI) metrics. A-OSE represents the total number of unknown objects misclassified as known objects. WI is calculated using the formula:
\begin{equation}
    WI = \frac{A-OSE}{TP_{\mathcal{K}} + FP_{\mathcal{K}}} 
\end{equation}
where $TP_{\mathcal{K}}$ and $FP_{\mathcal{K}}$ are the true positive and false positive predictions of known classes, respectively.

\subsubsection{Implementation Details.}

\noindent\textbf{Detailed Implementation of REW.} The encoder and decoder in REW are constructed using a single convolutional layer with $1\times 1$ kernels. We adopt the frozen ResNet50-FPN pre-trained by SoCo~\cite{wei2021aligning} as the backbone network for regional feature extraction. Since FPN is employed as the backbone, the multi-level feature maps corresponding to anchor boxes of different sizes exhibit significant variations in the number of region features. To prevent overfitting or underfitting the regional data in different levels, we build auto-encoders with different intermediate layer dimensions for the multi-level feature maps. For FPN feature maps $\{P_3,P_4,P_5,P_6\}$, the intermediate layer dimensions of auto-encoders are set to $\{32, 16, 8, 4\}$. The feature maps $\{P_3,P_4,P_5,P_6\}$ are corresponding to objects with sizes ranging from $[32^2, 64^2]$, $[64^2, 128^2]$, $[128^2, 256^2]$, $[256^2, 512^2]$ respectively, when calculating their reconstruction errors using Eq. \ref{equ::re}. Specifically, different Weibull models are formed for the region features at different levels as shown in Fig.~\ref{fig::rew_dists}, thereby enabling our REW model to effectively recognize unknown objects with varying sizes. We train REW using the MS-COCO training images, using \\ Eq. \ref{equ:rew} for 12 epochs. This training process does not require any object labels. For each task in OWOD, we model the Weibull distributions of the foreground and background based on the known object annotations, and subsequently assign soft labels to each pseudo label, as elaborated in Sec.~\ref{sec:weibull} and Sec.~\ref{sec:softlabel}.

\noindent\textbf{Detailed Implementation of Object Detector.} Our method is implemented based on Detectron2~\cite{wu2019detectron2}. We use ResNet-50~\cite{he2016deep} with FPN~\cite{lin2017feature} as the backbone network and Faster-RCNN\cite{ren2015faster} as the base detector. We adopt the SGD optimizer with an initial learning rate of 0.02, a momentum of 0.9, and a weight decay of $1e^{-4}$. All models are trained for 12 epochs under the standard schedule, and the self-training process lasts for 4 epochs for each round. All experiments are conducted using eight GPUs with a batch size of 16. We set the hyperparameters $\gamma =4$, $P =30$, and $l = 1$ by default in our experiments. 

\noindent\textbf{Pseudo Label Processing.}
We apply the same data processing strategy for the class-agnostic proposals generated by unsupervised region proposal generators and OLN to filter out low-quality pseudo labels. The proposals are first processed by Non-Maximum Suppression (NMS) at the IoU threshold of 0.3 to prevent overlapping bounding boxes. Then, only the proposals which satisfy the following requirements 
are used as pseudo labels for unknown objects: 
\begin{compactitem}
    \item The box size should be larger than 2000 pixels. 
    \item The aspect ratio should be between 0.25 and 4.0. 
    \item The IoU with known objects should be less than 0.3. 
\end{compactitem}
The first two rules aim to filter out low-quality boxes, while the purpose of the third rule is to prevent confusion with known instances.

\subsubsection{Competing Methods.}
Our method is compared with two baselines -- Faster-RCNN and Faster-RCNN + Finetuning -- and four state-of-the-art (SOTA) open-world object detectors, including ORE\cite{joseph2021towards}, OW-DETR\cite{gupta2022ow}, CAT\cite{ma2023cat} and PROB \cite{zohar2023prob}.
Faster-RCNN is only trained with labels of known classes. Faster-RCNN + Finetuning denotes that Faster-RCNN is finetuned using exemplar replay. They can only detect known objects, so their results of unknown recall are all zeros. 
The energy-based unknown identifier (EBUI) of ORE is not applied here since it needs weak unknown label supervision. 
The results of ORE, OW-DETR, CAT, and PROB are obtained from their paper. For the cross-dataset evaluation and label bias evaluation experiments, we also compete with the SOTA open-set object detection model, OpenDet\cite{han2022expanding}.

\subsection{Comparison to State-of-the-art}

\begin{table*}[t]
    \renewcommand\arraystretch{1.05}
    \centering
    \setlength{\tabcolsep}{3pt}
    \adjustbox{width=\textwidth}{
    \begin{tabular}{@{}l|ccccc|ccccc|ccccc@{}}
    \hline
        \textbf{Task IDs} ($\rightarrow$)& \multicolumn{5}{c|}{\textbf{Task 1}} & \multicolumn{5}{c|}{\textbf{Task 2}} & \multicolumn{5}{c}{\textbf{Task 3}}  \\ \hline
        & \multicolumn{3}{c} {Unknown Recall} & \multicolumn{2}{c|} {Confusion Metric}& \multicolumn{3}{c} {Unknown Recall} & \multicolumn{2}{c|} {Confusion Metric}& \multicolumn{3}{c} {Unknown Recall} & \multicolumn{2}{c} {Confusion Metric} \\
        
        &{R@10}&R@30&R@100&A-OSE&WI&R@10&R@30&R@100&A-OSE&WI&R@10&R@30&R@100&A-OSE&WI \\ 
        & ($\uparrow$)  & ($\uparrow$)  & ($\uparrow$) & ($\downarrow$) & ($\downarrow$) & ($\uparrow$)  & ($\uparrow$) & ($\uparrow$) & ($\downarrow$) & ($\downarrow$) & ($\uparrow$)  & ($\uparrow$) & ($\uparrow$) & ($\downarrow$) & ($\downarrow$) \\ \hline

\begin{tabular}[c]{@{}l@{}}Faster-RCNN\\ \hspace{1em}+ Finetuning\end{tabular} & 0.0  & 0.0  & 0.0 & 1807 & 0.022 & 0.0 & 0.0 & 0.0 & 4007 & 0.033 & 0.0  & 0.0  & 0.0 & 4010 & 0.025\\ \hline

ORE $-$ EBUI~\cite{joseph2021towards}& 3.6  & 6.3 & 11.9 & 2486 & 0.024 & 5.6  & 10.5 & 14.5 & 6608 & 0.040 & 6.8 & 10.7 & 13.1 & 6896 & 0.026 \\

OW-DETR~\cite{gupta2022ow} &  2.3  &  5.6 &  14.7 &  12721 &  0.029 &  4.3 & 9.0 & 19.2 & 14970 & 0.041 & 4.4 & 13.5 & 24.2 & 9197 & 0.024 \\ 

{PROB~\cite{zohar2023prob}}&11.6&23.2&37.8& 2003 & 0.021&14.6&26.8&40.5 &3358&0.031 &15.2 & 27.1 &43.5&\textbf{1546}&\textbf{0.018}\\ 

{CAT~\cite{ma2023cat}} &13.5& 26.1& 40.5& 2097& 0.023& 13.3& 25.6& 40.1&5784& 0.040& 14.1& 26.5& 42.7 & 3545 & 0.021\\

\textbf{MEPU-FS (Ours)} &\textbf{25.1} &\textbf{39.4} &\textbf{54.8} &\textbf{1710} & \textbf{0.020}&\textbf{25.0} &\textbf{40.1} &\textbf{54.4} &\textbf{3197} &\textbf{0.027} & \textbf{24.8}&\textbf{39.7} & \textbf{55.5} & {2862} & {0.020} \\

\textbf{MEPU-SS (Ours)} &21.2&36.2&54.5 & 1753&\textbf{0.020} & 22.3& 36.6& 53.1& 3352& 0.028& 23.3& 36.5& 52.7 & 2883 & {0.020}\\

\bottomrule
\toprule 

\begin{tabular}[c]{@{}l@{}}
Faster-RCNN\\ \hspace{1em}+ Finetuning\end{tabular} & 0.0  & 0.0  & 0.0 &13396& 0.070 & 0.0  & 0.0 & 0.0 &  12291 & 0.037 & 0.0  & 0.0 & 0.0 & 9622 & 0.028 \\ \hline

ORE $-$ EBUI~\cite{joseph2021towards}& 3.8  & 9.4 & 17.5 & 10459& 0.062&  4.4  & 9.8 & 15.3 & 10445&0.028 & 4.2 & 8.5 & 16.8 & 7990&0.021 \\

OW-DETR~\cite{gupta2022ow} & 6.2  & 12.3 & 19.9 &10240&{0.057}  & 5.1  & 11.3 & 19.1 & 8441 & 0.028 & 6.9 & 12.3 & 18.5 &6803 &{0.016}  \\ 

{PROB~\cite{zohar2023prob}} &10.7&19.8 &32.2&\textbf{5195} &0.057 &12.3 &17.8 &32.5 &6452 &0.034 &11.2 &18.1 &33.0&\textbf{2641}&\textbf{0.015} \\

{CAT~\cite{ma2023cat}} &11.8& 20.5& 34.0& 20364& 0.066& 12.1&18.7&33.2&16768&0.032& 12.9&19.5& 33.7& 7515&0.020 \\

{\textbf{MEPU-FS (Ours)}} &\textbf{18.9}&\textbf{27.6} & \textbf{38.9}&{6050} & \textbf{0.056}& \textbf{18.4}&\textbf{26.1}& \textbf{37.5} &\textbf{5925} &\textbf{0.023} &\textbf{19.2} &\textbf{27.6} &\textbf{36.9}  &{5159}  &{0.016}  \\

{\textbf{MEPU-SS (Ours)}} &16.7& 25.5& 38.5& 6235 & 0.057& {17.0}&{24.3} & {37.0}& 6042&\textbf{0.023} & 18.8& 26.3& 36.2 & 5369 & {0.016}\\

    \bottomrule
    \end{tabular}%
    }
     \caption{ \textbf{Additional results for OWOD on S-OWODB (top) and M-OWODB (bottom).} We report the top K recall of unknown objects (noted as R@K) and the confusion metrics (including A-OSE and WI) for a more detailed analysis.  }
    \label{tab:additional_owod}
    
    \end{table*}

\subsubsection{Open-World Object Detection Performance}
The mean Average Precision (mAP) for known classes and U-Recall for unknown classes of our method and state-of-the-art models on S-OWODB and M-OWODB are presented in Tab.~\ref{tab:sota_owod}.
The results of our model using FreeSOLO \cite{wang2022freesolo} and Selective Search \cite{uijlings2013selective} as proposal generators, denoted as MEPU-FS and MEPU-SS repetitively, are reported here. 

Overall, both variants of our method, MEPU-FS, and MEPU-SS, consistently and substantially outperform the four SOTA open-world methods in terms of both unknown and known object detection across the four tasks, while retaining almost the same detection performance as the closed-set baseline model Faster-RCNN + Finetuning on detecting the known objects. Detailed analyses of the results are presented separately as follows.

\noindent\textbf{Unknown Object Detection.} Our method, MEPU-FS, exhibits significantly better performance in detecting unknown objects compared to the top competitor CAT. On S-OWODB, MEPU-FS surpasses CAT by 13.9 (Task 1), 12.8 (Task 2), and 11.1 (Task 3) in U-Recall. MEPU-SS, another variant of our method using the traditional proposal generator, also achieves notable improvements in U-Recall performance, with gains of 9.3 (Task 1), 11.2 (Task 2), and 9.0 (Task 3) over CAT. As for M-OWODB, our MEPU-FS model achieves gains of 7.9 (Task 1), 11.8 (Task 2), and 5.7 (Task 3) in U-Recall due to our strong capability to overcome the label bias problem. Furthermore, as shown in Tab.~\ref{tab:additional_owod}, our MEPU-FS outperforms CAT and PROB in Unknown R@10 by 6.3 - 13.5, which indicates that our model can identify unknown objects with higher accuracy. MEPU-FS also largely improves Unknown R@100 by 3.2 - 17.0, demonstrating our advantage in retrieving unknown objects. Besides, we also achieve SOTA performance in terms of WI and A-OSE on S-OWODB Task 1-2 and M-OWODB Task 2, which shows that our method can effectively reduce confusion between unknown and known classes. 

\noindent \textbf{Known Object Detection.} The four competing methods sacrifice the close-set performance for the open-set detection, leading to large performance drops in Known-mAP across four tasks, \eg, maximally a decrease of 7.5 for ORE, 6.3 for OW-DETR, 3.9 for CAT, and 4.2 for PROB on S-OWODB. By contrast, our methods have nearly the same, or even better, close-set detection ability as the closed-set baseline, Faster-RCNN + Finetuning, on all four tasks. The consistently superior performance of our methods across the tasks is mainly because our unsupervised unknown object modeling is able to generate accurate pseudo labels. 

\noindent \textbf{Incremental Detection.} In task-incremental learning, MEPU-FS demonstrates superior performance compared to ORE and OW-DETR in detecting both previously and currently known objects. It achieves improvements ranging from 1.2 to 6.7 in previously-known-mAP and 7.0 to 12.6 in currently-known-mAP. Although our model may be relatively less effective on M-OWODB Tasks 3-4, it still outperforms CAT and PROB on other tasks. These results highlight the substantially enhanced ability of our methods to acquire new knowledge without catastrophically forgetting previously obtained knowledge.

\subsubsection{Cross-dataset Generalization} 
To evaluate the generalization ability for large datasets which is closer to real-world situations, the models are trained on COCO \texttt{train2017} with 80 known classes, and evaluated on the validation set of LVIS and Objects365. 
The classes on LVIS and Objects365 beyond the 80 COCO classes are treated as unknown object classes. The results of our method MEPU-FS, MEPU-FS, the baseline Faster-RCNN, SOTA OSOD model OpenDet\cite{han2022expanding}, and SOTA OWOD model CAT are reported in Tab.~\ref{tab:cross_dataset}. Our method MEPU-FS outperforms OpneDet in detecting unknown objects for both datasets in terms of R@10, R@30, and R@100, which shows our superiority in recognizing cross-dataset unknown objects. Further, MEPU-FS substantially outperforms OpneDet in Known-mAP by 2.3 on LVIS and 1.7 on Objects365, indicating a much stronger generalization ability in detecting known objects under the cross-dataset setting.

\begin{table} 
    \renewcommand\arraystretch{1.05} 
    \centering 
    \footnotesize 
    \adjustbox{width=0.48\textwidth} { 
        \begin{tabular}{l|c|c|ccc} 
        \hline 
        \makebox[0.06\textwidth][l]{\multirow{2}*{Method}}&  \makebox[0.05\textwidth][c] {\multirow{2}*{Data}}  &  \makebox[0.05\textwidth][c]{Known} & \multicolumn{3}{c} {Unknown}  \\ 
        &&mAP&\makebox[0.03\textwidth][c]{R@10}&\makebox[0.03\textwidth][c]{R@30}&\makebox[0.04\textwidth][c]{R@100} \\ \hline 
        Faster-RCNN&\multirow{5}*{\makecell{LVIS}}&\textbf{39.3}&0.0&0.0&0.0  \\ 
        CAT&&35.5&13.3&23.7&36.5   \\ 
        OpenDet&&35.2&14.1&26.8&41.8   \\  
        \textbf{MEPU-SS (Ours)}&&37.0&{15.0}&{29.5}&{45.3}  \\
        \textbf{MEPU-FS (Ours)}&&37.5&\textbf{16.2}&\textbf{30.1}&\textbf{45.6}  \\ \hline
        Faster-RCNN&\multirow{5}*{\makecell{Objects-\\365}}&\textbf{38.2}&0.0&0.0&0.0  \\ 
        CAT&&35.1&11.9&21.7&38.8   \\ 
        OpenDet&&34.0&13.6&26.9&42.0   \\  
        \textbf{MEPU-SS (Ours)}&&35.5&{16.5}&{30.1}&{45.8}  \\
        \textbf{MEPU-FS (Ours)}&&35.7&\textbf{17.2}&\textbf{30.5}&\textbf{46.0}  \\ \hline
        \end{tabular}
    }
    
    \caption{\textbf{Results of cross-dataset generalization.}}
    
    \label{tab:cross_dataset}
    \end{table}

\begin{table*}[t]
\renewcommand\arraystretch{1.2}
    \centering
    \normalsize
    \scalebox{0.9}{
    \begin{tabular}{@{}l|ccc@{}} 
    \hline 
        & \textbf{Known} & \textbf{Related Unknown} & \textbf{Unrelated Unknown} \\ \hline
    Semantic split & \begin{tabular}[c]{@{}c@{}}Animals(5), Person(1), \\ Vehicles(4), Outdoor(3), \\ Furniture(3), Appliances(2), \\ Accessories(2)  \end{tabular} & \begin{tabular}[c]{@{}c@{}}Animals(5), Vehicles(4), \\  Outdoor(2), Furniture(3), \\ Appliances(3), Accessories(3) \end{tabular} &  
    \begin{tabular}[c]{@{}c@{}}Sports(10), Food(10), \\ Electronic(6), Indoor(7), \\ Kitchen(7)\end{tabular} \\
    \hline
    \end{tabular}%
    }
    \caption{\textbf{Proposed dataset split of MS COCO for evaluating the impact of label bias problem. } 
        Numbers in brackets represent the number of classes belonging to each super-category.}
    \label{tab:data_split_label_bias}
    \end{table*}

\begin{table*}
    \centering
    \renewcommand\arraystretch{1.2}
    \footnotesize
    \adjustbox{width=1.0\textwidth}{
        \begin{tabular}{c|c|ccc|ccc}
        \hline
        & {Known} &  \multicolumn{3}{c|}{Related Unknown} &  \multicolumn{3}{c}{Unrelated Unknown}  \\ 
        &mAP&{R@10}&{R@30}&{R@100}&{R@10}&{R@30}&{R@100}  \\ \hline
        CAT&63.8&25.1&41.2&52.1&11.6&23.3&39.1  \\ \hline
        OpenDet&62.2&27.0&43.5&59.5&13.7&26.5&45.5  \\ \hline
        \textbf{MEPU-FS (Ours)}&\textbf{66.9(+4.7)}&\textbf{27.8(+0.8)}&\textbf{45.2(+1.7)}&\textbf{60.9(+1.4)}&\textbf{25.0(+11.3)}&\textbf{37.3(+10.8)}&\textbf{51.1(+5.6)}\\ \hline
        \textbf{MEPU-SS (Ours)}&66.7(+4.5)&27.2(+0.2)&44.6(+1.1)&60.5(+1.0)&21.3(+7.6)&36.0(+9.5)&51.0(+5.5) \\ \hline
        \end{tabular} 
    }
    
    \caption{\textbf{Results on the proposed dataset split for handling the label bias issue. } }
    \label{tab:label_bias}
    \end{table*}
    
\subsubsection{Label Bias Evaluation}
To further analyze the label bias problem discussed in Sec.~\ref{sec:intro}, we propose a new dataset split that divides COCO into three parts: Known, Related Unknown, and Unrelated Unknown, as shown in Tab.~\ref{tab:data_split_label_bias}.  Classes within `Related Unknown' have semantic relations with known ones. For example, there are five known animal classes and another five that belong to the `Related Unknown' category. The `Unrelated Unknown' classes belong to separate super-categories from the `Known' ones. The models are trained only using the known object annotations in COCO \texttt{train2017}, and tested on COCO \texttt{val2017} to evaluate the detection performance of known, related unknown, and unrelated unknown classes. 

The evaluation results of our MEPU-FS, MEPU-SS, the most competitive method OpenDet, and CAT on the proposed split are demonstrated in Tab.~\ref{tab:label_bias}. All models are trained using the known categories on MS-COCO, and tested on the known, related unknown, and unrelated unknown classes of COCO \texttt{val2017} respectively. It can be observed that OpenDet can achieve similar performance as MEPU-FS in the `Related Unknown' classes (\eg, 27.0 vs 27.8 in R@10), but it fails to detect the unknown objects which have no semantic relations to the known ones due to its severe label bias problem. Our MEPU-FS can effectively eliminate the label bias (\eg, + 11.3 Unrelated Unknown R@10) because we recognize unknown objects based on the shared region feature frequencies between the known and unknown classes, which can largely improve our generalization ability toward unknown objects, especially the unknown objects that have no semantic relations with the known classes.

\subsection{Ablation Study\label{sec:ablation}}

\begin{table}
    \renewcommand\arraystretch{1.05}
    \centering
    \footnotesize
    \adjustbox{width=0.48\textwidth} {
    \begin{tabular}{ccc|c|ccc}
    \hline
    {\multirow{2}*{REW}}  & {\multirow{2}*{ROLNet}}  & {{Proposal} }& {Known}&  \multicolumn{3}{c}{Unknown}    \\ 
    &&Generator&mAP&\makebox[0.04\textwidth][c]{R@10}&\makebox[0.04\textwidth][c]{R@30}&\makebox[0.04\textwidth][c]{R@100} \\ \hline
    \XSolidBrush &\XSolidBrush&FS&71.8&22.1&31.5&45.2  \\ 
    \Checkmark &\XSolidBrush&FS&74.1&24.2&36.6&50.3  \\ 
    \XSolidBrush&\Checkmark &FS&72.5&21.9&35.5&51.8  \\ 
    \rowcolor{gray!40} \Checkmark &\Checkmark  &FS&\textbf{74.3}&\textbf{25.1}& \textbf{39.4}& \textbf{54.8}  \\ \hline
    \XSolidBrush&\XSolidBrush&SS&70.6&11.0&17.8&34.8  \\ 
    \Checkmark &\XSolidBrush&SS&74.0&19.5&30.9&49.1  \\ 
    \XSolidBrush&\Checkmark &SS&71.4&15.6&32.4&50.0  \\ 
    \rowcolor{gray!40} \Checkmark &\Checkmark &SS&\textbf{74.2}&\textbf{21.2}&\textbf{36.2}&\textbf{54.5}  \\ \hline
    \end{tabular}%
    }

    \caption{\textbf{Our full model (shaded) vs. its variants.}}

    \label{tab:component_ablations}
    \end{table}
    
\noindent\textbf{The REW and ROLNet Modules. } We analyze the contribution of two main modules, REW and ROLNet, on the COCO dataset. The results of Known-mAP, Unknown R@10, R@30, and R@100 on the S-OWODB Task 1 are reported in Tab.~\ref{tab:component_ablations}. Without both components, the method is equivalent to directly training Faster-RCNN using 
proposals generated by unsupervised proposal methods as unknown pseudo labels. Such unknown pseudo labels are mostly inaccurate, making the detector
confused with foreground and background regions, which harms both closed-set and open-set performances. REW improves the Known-mAP 
significantly (+2.5 using FS and +3.6 using SS) and also contributes to unknown-recall performance, especially Unknown R@10 (+2.1 using FS and +8.5 using SS),
mainly because it helps recognize true unknown objects from the noisy pseudo labels. ROLNet aims to extend the incomplete pseudo unknown labels, so it mainly improves the Unknown-R@100 (+6.6 using FS and +15.2 using SS). By combining REW and ROLNet, we achieve our full model which effectively recognizes true unknown objects and accurately extends the incomplete unknown labels. This further improves the overall detection performance.

\begin{figure*}[t]
    \centering
        \centering
        \includegraphics[width=0.95\textwidth]{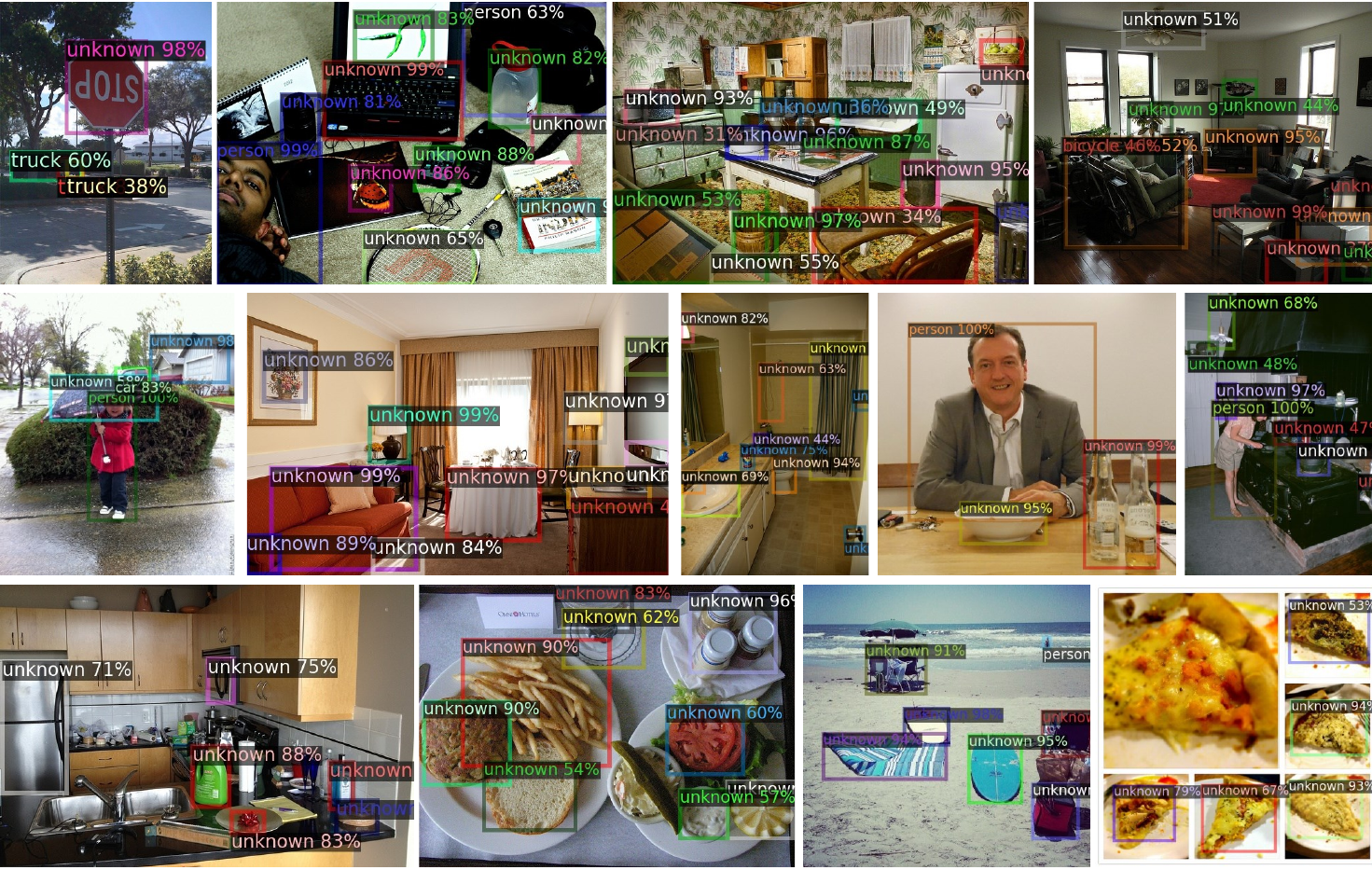}
        \caption{
            \textbf{Qualitative results of MEPU-FS on S-OWODB Task-1.} We sample the images from COCO \texttt{val2017} and plot the object predictions with scores higher than 0.3. Even though the known classes include persons, animals, and vehicles, our model can still recognize unknown objects, \eg, computer, chairs, and bottles, that do not have any semantic relation to the known ones.  
        }
        \label{fig:add_good}
    \end{figure*}

\noindent\textbf{Importance of Region Proposal Generators. }
We also evaluate the contribution of the unsupervised proposal generators to our model. As reported in Tab.~\ref{tab:region_proposal}, five unsupervised proposal generation methods are used and compared with a baseline method that does not use any proposal generators. The five methods include deep learning-based method DETReg~\cite{bar2022detreg} and learning-free methods (EdgeBoxes~\cite{zitnick2014edge} and Geodesic Object Proposals (GOP)~\cite{krahenbuhl2014geodesic}), in addition to FreeSOLO and Selective Search. All proposal generators generally produce better results than the baseline, enabling substantially improved unknown object detection results, especially R@10 (\eg, +8.8 using FreeSOLO). Note that although the traditional methods, such as Selective Search and GOP, mostly generate coarse bounding boxes with poor localization quality, they still improve the final performance since our REW module can recognize true unknown objects from the noisy labels.

\begin{figure*}[t]
    \centering
        \includegraphics[width=0.95\textwidth]{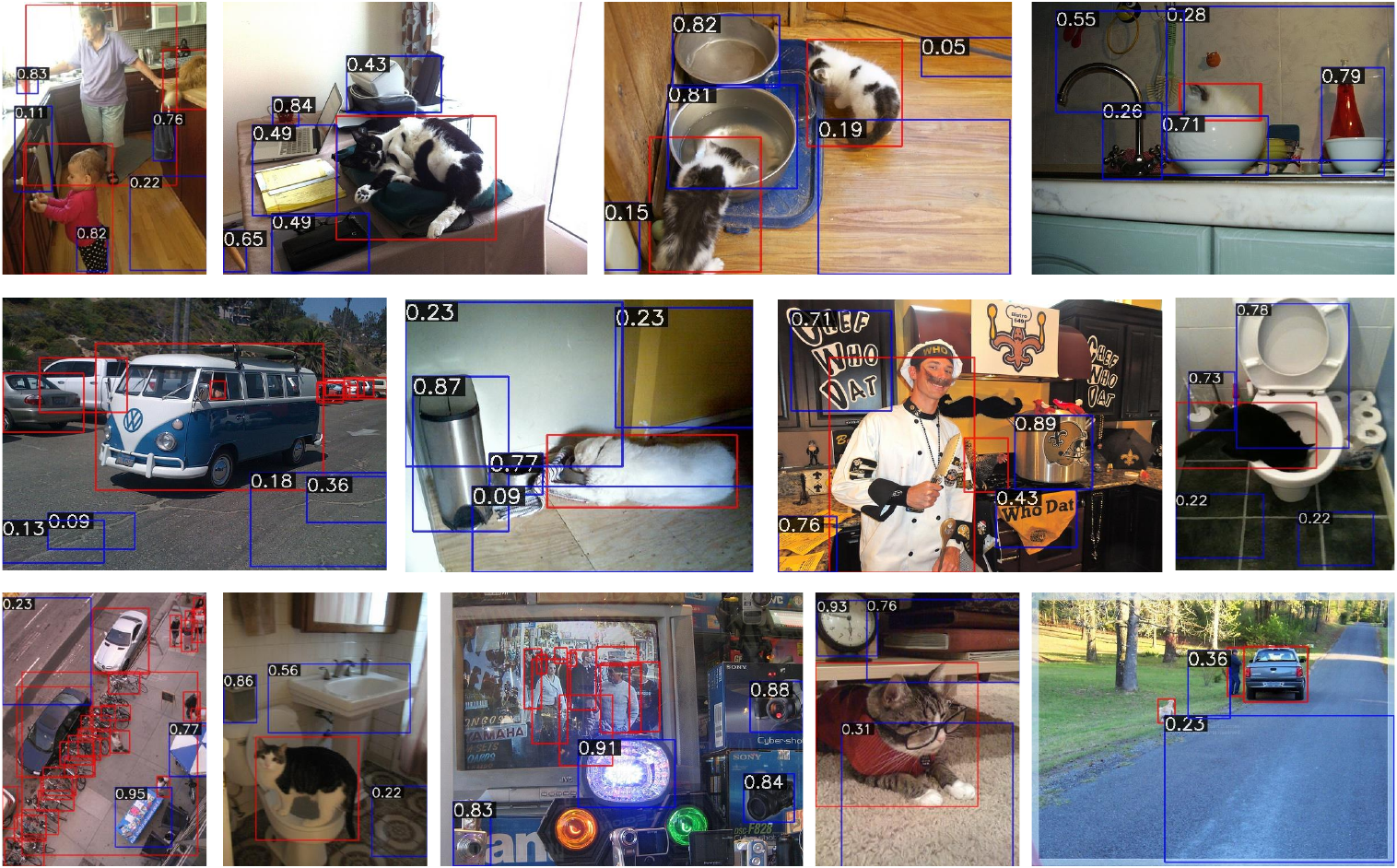}
        \caption{
            \textbf{Visualizations of our REW module on M-OWODB Task-1.} The blue bounding boxes denote the pseudo labels generated by FreeSOLO on COCO-Task 1. The numbers on the top left corner of the blue boxes indicate the soft labels yielded by REW for the pseudo unknown objects. The red bounding boxes denote the known object labels.
        }
        \label{fig:vis_rew}
    \end{figure*}

\begin{table}[t]
    \renewcommand\arraystretch{1.05}
    \centering
    \footnotesize
    \begin{tabular}{@{}l|c|ccc@{}}
    \hline
    \makebox[0.15\textwidth][l]{\multirow{2}*{Proposal Generator}} & \makebox[0.05\textwidth][c]{Known} & \multicolumn{3}{c}{Unknown}   \\ 
    &mAP&\makebox[0.04\textwidth][c]{R@10}&\makebox[0.04\textwidth][c]{R@30}&\makebox[0.05\textwidth][c]{R@100} \\ \hline
    Baseline &73.5&16.3&31.2&48.2 \\ 
    DETReg~\cite{bar2022detreg}&74.2&22.5&37.2&53.5  \\  
    EdgeBoxes~\cite{zitnick2014edge}&73.6&21.5&36.5&52.7  \\  
    GOP~\cite{krahenbuhl2014geodesic}&73.8&19.3&35.1&52.0  \\
    Selective Search~\cite{uijlings2013selective}&74.2&21.2&36.2&54.5  \\ 
    FreeSOLO~\cite{wang2022freesolo}&74.3&25.1& 39.4&54.8  \\ \hline
    \end{tabular}%
    \caption{\textbf{Results of using different unsupervised region proposal methods to generate pseudo labels.} `Baseline' is a method trained using only self-training without region proposal generators.}
    \label{tab:region_proposal}
\end{table}

\noindent\textbf{Analysis of Hyperparameters. }
Tab.~\ref{tab:para_ablations} shows the impact of using different settings of three key hyperparameters on S-OWODB Task-1. $\gamma$ controls the scores of the soft labels in REW. When adopting a small $\gamma$, it cannot effectively reduce the negative impact from the background regions; whereas a large $\gamma$ can lead to the rejection of most pseudo labels. It can be observed that $\gamma = 4.0$ produces balanced results. $P$ is to retain the top $P$\% proposals as pseudo labels for self-training in ROLNet, so it affects the quality of pseudo labels. 
We find that $P = 30$ achieves a good trade-off between the unknown recalls. 
$l$ is the number of self-training iterations in ROLNet. There is a significant performance gain compared with no self-training (\ie, $l=0$), especially R@100, since ROLNet can extend the incomplete pseudo labels. While self-training for more than one iteration leads to a drop in R@10, because the extended labels can be less accurate (\eg, with poorer localization quality).

\begin{table}[t]
    \renewcommand\arraystretch{1.05}
    \centering
    \footnotesize
    \begin{tabular}{ccc|c|ccc}
    \hline
    {\multirow{2}*{$\gamma $}}&{\multirow{2}*{$P$}} &{\multirow{2}*{$l$}} & {Known} &  \multicolumn{3}{c}{Unknown}   \\ 
    &&&mAP&{R@10}&{R@30}&{R@100} \\\hline
    0.5&30&1&73.4&23.0&36.0&52.3  \\ 
    1.0&30&1&73.8&24.0&38.2&54.5  \\ 
    2.0&30&1&73.8&24.8&38.5&54.2  \\  
    4.0&30&1&\textbf{74.3}&25.1&\textbf{39.4} &54.8  \\ 
    8.0&30&1&74.2&\textbf{25.2}&38.5&52.8  \\\hline
    4.0&10&1&74.1&\textbf{25.2}&38.6&52.3 \\
    4.0&50&1&74.0&22.4&36.8&53.6 \\
    4.0&70&1&73.8&21.0&35.3&53.0 \\\hline
    4.0&30&0&74.1&24.2&35.6&48.3 \\
    4.0&30&2&74.2&24.4&38.5&\textbf{55.4} \\ 
    4.0&30&3&74.0&24.0&38.7&55.0 \\\hline
    \end{tabular}%
    \caption{\textbf{Analysis on hyperparameter settings.}}
    \label{tab:para_ablations}
\end{table}

\noindent\textbf{Reconstruction error-based OOD detection vs. REW.}
Previous reconstruction-based OOD detection models typically adopt reconstruction error as OOD scores and reject all samples whose scores are higher than a pre-define threshold during testing. The comparison results of our REW model (soft labeling) with the base model (original reconstruction error-based OOD detection) are shown in Tab.~\ref{tab:hard_thr}. The threshold in the OOD detector is chosen at 95\% true positive rate (foreground is the positive category). Our REW surpasses the base model in both known mAP and unknown recall, since (i) some true unknown objects may be filtered out by the non-adaptive hard thresholding in the base model, leading to high false negative errors, (ii) whereas the Weibull modeling in REW provides an adaptive recognition of unknown objects from the background regions, achieving substantially enhanced unknown object recognition performance. 

\noindent\textbf{Alternative Designs in REW and ROLNet.} We also evaluate different variants of REW and ROLNet on S-OWODB Task-1, with the results reported in Tab.~\ref{tab:option_ablations}. For REW, we test the effectiveness of replacing its offline training with an online training strategy, meaning we train the auto-encoder together with the detection model using the same backbone, rather than a separate reconstruction backbone; as for ROLNet, one alternative approach is to produce the pseudo labels directly using the predictions of RPN instead of OLN. The results show that, compared to our default model (MEPU-FS), using the online training in REW leads to a performance drop (-0.3 Known mAP and -4.1 Unknown R@100) because the downstream tasks in object detection affect the semantic expressiveness of dense features (\eg, it is biased towards features of known objects), which is also discussed in \cite{liu2022open, winkens2020contrastive}. When self-training without employing OLN, the Unknown R@100 dropped by 6.1. This decrease can be attributed to the fact that the classification loss of Faster-RCNN heavily suppressed the scores of unlabeled regions, making it challenging to extend the detection to unlabeled objects.

\begin{table}[t]
    \centering
    \renewcommand\arraystretch{1.1}
    \footnotesize
    \adjustbox{width=0.48\textwidth}{
        \begin{tabular}{c|c|c|ccc}
        \hline
        {\multirow{2}*{Method}}  &{Proposal}& {Known} &  \multicolumn{3}{c}{Unknown}  \\ 
        &Genrator&mAP&{R@10}&{R@30}&{R@100}  \\ \hline
        
        {Base Model} &FS&72.9&23.2&33.9&45.6  \\
        (Hard Threshold)&SS&71.6&15.2&27.3&43.1  \\  \hline
        REW&FS&74.1&24.2&36.6&50.3 \\
        (Soft Labeling)&SS&74.0&19.5&30.9&49.1  \\ \hline
        \end{tabular}%
    }

    \caption{\textbf{Our REW model (soft labeling) vs. base model (hard threshold).} }
    \label{tab:hard_thr}
    \end{table}

    

\begin{table}[t]
    \centering
    \renewcommand\arraystretch{1.1}
    \footnotesize
    \adjustbox{width=0.48\textwidth}{
        \begin{tabular}{c|c|ccc}
        \hline
        {\multirow{2}*{Design}} & {Known} &  \multicolumn{3}{c}{Unknown}  \\ 
        &mAP&{R@10}&{R@30}&{R@100}  \\ \hline
       Online Training&74.0&23.5&36.4&50.7  \\  \hline
        \begin{tabular}[c]{c}Self-Training   \\w/ RPN\end{tabular}&74.2&23.9&35.6&48.7  \\ \hline
       Our Default Model &\textbf{74.3}&\textbf{25.1}&\textbf{39.4} &\textbf{54.8}  \\ \hline
        \end{tabular}%
    }
    
    \caption{\textbf{Comparison to alternative REW and ROLNet designs.} Either online training in REW or RPN-based self-training in ROLNet is used. }
    \label{tab:option_ablations}
    \end{table}

\subsection{Qualitative Results.\label{sec:qualitative}}

\begin{figure*}
    \centering
        \centering
        \includegraphics[width=.95\textwidth]{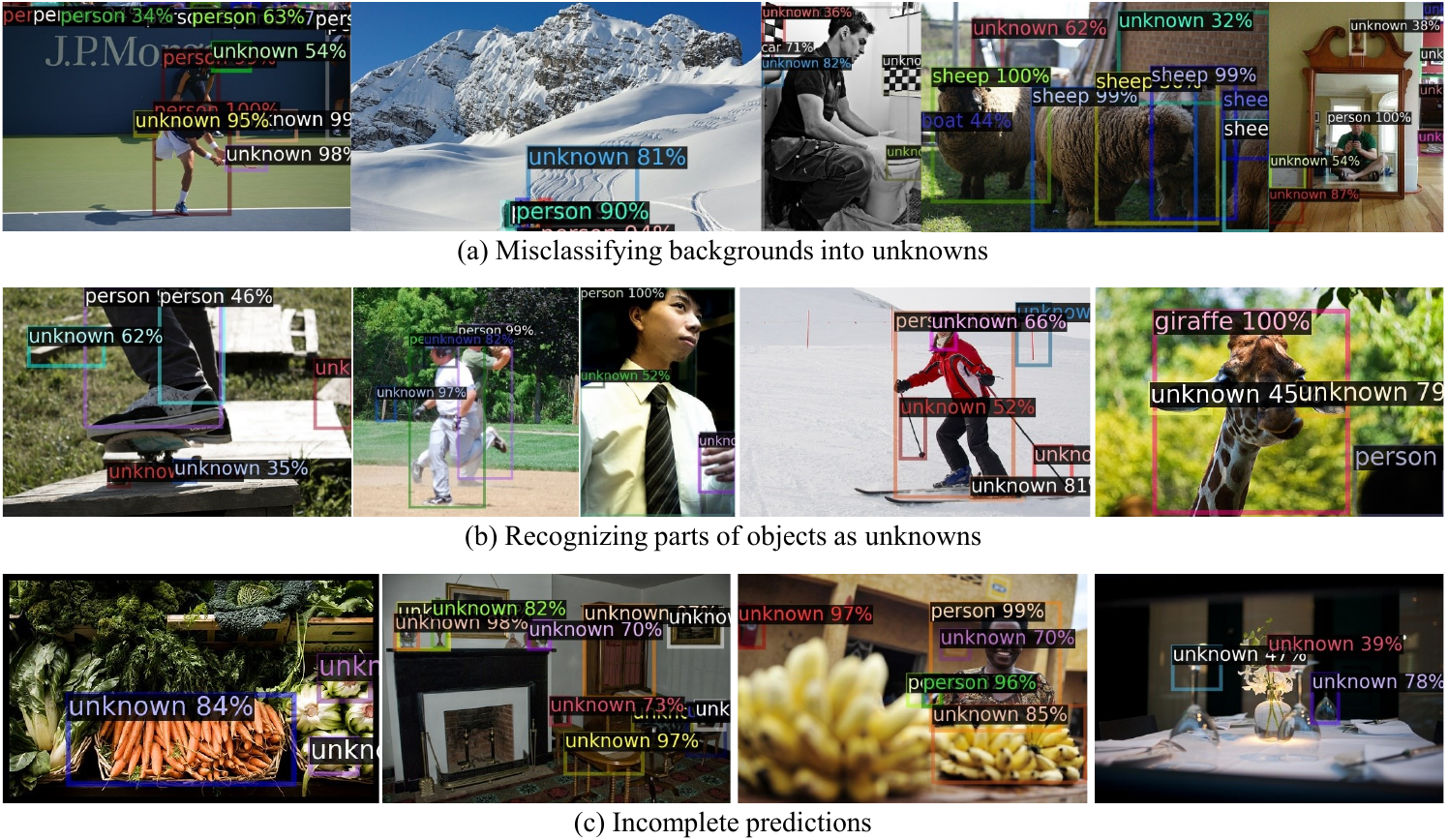}
        \caption{
            \textbf{Failure cases of our method on S-OWODB Task-1.}
        }
        \label{fig:add_fail}
    \end{figure*}
    
\noindent\textbf{Visualizations of MEPU-FS.} Fig.~\ref{fig:add_good} visualizes the object predictions of our full model MEPU-FS on S-OWODB Task-1. Although the known classes only include persons, animals, and vehicles, our method can still recognize various types of unknown objects, \eg, computers, chairs, and sofas, which do not have any semantic relation to known ones. This remarkable capability can be attributed primarily to our REW module, which autonomously identifies unknown objects, ultimately mitigating the label bias problem.

\noindent\textbf{Effectiveness of REW.} We further conducted a qualitative study of our REW module to provide a more detailed analysis. Fig.~\ref{fig:vis_rew} visualizes the soft labels generated by REW on S-OWODB Task-1, indicating the likelihood score of a pseudo object being a true unknown object. For true unknown object proposals or boxes covering both foreground and background regions, REW assigns relatively higher scores (e.g., 0.7 - 1.0) or medium scores (e.g., 0.3 - 0.7). In contrast, for non-object background regions such as the sky, road surface, floor, and white wall, REW produces lower scores (e.g., 0.0 - 0.3). These largely help enhance the accuracy of unknown object recognition by our detector. Notably, REW can assign high scores for unknown objects that do not exhibit similar appearances to known objects, since it utilizes the shared regional feature frequencies to recognize unknown objects. Instead, previous methods like OWOD/OSOD rely on objectness scores based on the supervision of known objects for unknown object pseudo-labeling. Consequently, they suffer from a severe label bias problem towards the known classes, which reduces their generalization ability.

\noindent\textbf{Failure Cases.} Although our model exhibits superior performance in identifying unknown objects compared to previous methods, there are inevitably cases of failure in certain situations, since we do not leverage any labels of unknown classes. Fig.~\ref{fig:add_fail} illustrates typical failure cases of MEPU, such as misclassifying backgrounds with complex patterns as unknowns, identifying parts of objects as unknowns, or producing incomplete predictions.

\section{Conclusion}

\noindent This work proposes the MEPU approach that addresses the label bias problem in OWOD by identifying unknown objects in an unsupervised manner and iteratively extending the identified pseudo unknown objects. To this end, it first performs an unsupervised recognition of unknown objects from the pseudo labels generated by unsupervised region proposal methods, and then extends new unknown objects that are not covered by the pseudo labels. These two tasks are implemented by two main components, REW and ROLNet. Experiments show that our MEPU approach achieves superior performance in detecting unknown objects on S-OWODB and M-OWODB. Besides, our method also outperforms existing models on cross-dataset experiments on the LVIS and Object365 datasets. Although the detection performance gap between known objects and unknown ones is still large, our work makes a step forward in OWOD for real-world applications.

\bibliographystyle{IEEEtran}
\bibliography{reference}


\end{document}